%% file: main.tex
\pdfoutput=1
\documentclass[11pt, a4paper, copyright]{perplexity}

\usepackage[authoryear, sort&compress, round]{natbib}
\input{extrapackages}

\renewcommand{\copyrightext}{\footerfont \textcopyright\, \the\year{} Perplexity and turbopuffer. All rights reserved}

\uselogo{}

\title{Contextual Embeddings Beyond the Gold Chunk}

\author[*,1]{Sedigheh Eslami}
\author[*,1]{Markus Krimmel}
\author[2]{William Barber}
\author[1]{Antoine Chaffin}
\author[1]{Maximilian Schall\textsuperscript{\textdagger}}
\author[1]{Jonas Golde}
\author[1]{Louis Milliken}
\author[1]{Ekaterina Glazkova\textsuperscript{\textdagger}}
\author[1]{Denis Bykov}
\author[1]{Bo Wang}

\affil[1]{Perplexity AI}
\affil[2]{turbopuffer}
\affil[*]{Equal contribution, \textsuperscript{\textdagger}Work done while at Perplexity}

\begin{abstract}
Contextual embeddings represent chunks of a document with respect to the document's global context. These models are typically trained and evaluated under the assumption that a relevant document contains only a single relevant chunk. In this work, we argue that in real-world use cases a retriever should not only find answers spread over several chunks but also surface the evidence needed to verify them. To address this, we introduce \vtwo{}, a family of contextual embedding models trained by distilling graded chunk-level relevance from a query-aware context compression model. This removes the need for gold-chunk labels and makes the training targets independent of the chunking. We utilize a vision-language backbone for training and release \vtwo{} in 9B and 0.6B sizes with compatible embedding spaces. On the evaluation side, we introduce two benchmarks that label both the chunks that answer a query and the evidence needed to verify the answer. Context-bench is a controlled diagnostic, and \qdchunk{} is built from real-world search queries. To encourage further research, we also publish \qdchunk{} as a public leaderboard. Our 9B model sets a new state of the art on both benchmarks and ConTEB.
We find that the visual retrieval capabilities of our 9B model transfer from its vision-language base model and propose to extend chunking to image inputs. We demonstrate that our models achieve state-of-the-art performance among dense retrievers on visual ViDoRe-V3 and that performance improves with granular image chunking.
\end{abstract}

\begin{document}

\maketitle

\input{core/1_intro}
\input{core/3_method}

\input{core/4_benchmark}
\input{core/5_results}
\input{core/6_ablations}
\input{core/7_conclusion}

\bibliography{main}

\clearpage
\input{core/8_appendix}

\end{document}

%% file: extrapackages.tex
\usepackage{graphicx}
\usepackage{multirow}
\usepackage{float}
\usepackage{adjustbox}
\usepackage{tablefootnote}
\usepackage[bottom]{footmisc}
\usepackage{subcaption}
\usepackage{tcolorbox}
\usepackage{tikz}
\usetikzlibrary{positioning,arrows.meta,decorations.pathreplacing}
\tcbuselibrary{breakable}

\providecommand{\doi}[1]{\href{https://doi.org/#1}{doi:\,\nolinkurl{#1}}}

\newtcolorbox{todobox}[1][TODO]{
  breakable, colback=red!6, colframe=red!70!black, boxrule=0.8pt, arc=2pt,
  left=6pt, right=6pt, top=4pt, bottom=4pt,
  fonttitle=\bfseries\small, title={#1}, coltitle=white,
}

\providecommand{\code}[1]{\texttt{#1}}
\providecommand{\vone}{\texttt{pplx-\allowbreak embed-\allowbreak context-\allowbreak v1}}
\providecommand{\vonesize}[1]{\texttt{pplx-\allowbreak embed-\allowbreak context-\allowbreak v1-#1}}
\providecommand{\vtwo}{\texttt{pplx-\allowbreak embed-\allowbreak v2-\allowbreak context}}
\providecommand{\vtwosize}[1]{\texttt{pplx-\allowbreak embed-\allowbreak v2-\allowbreak context-#1}}
\newcommand{\contextbench}{context-bench}
\newcommand{\Contextbench}{Context-bench}
\newcommand{\qdchunk}{Q2D-Chunk}

\definecolor{pcanswer}{HTML}{F2C200}
\definecolor{pcevidence}{HTML}{339AF0}
\definecolor{pcother}{HTML}{ADB5BD}
\newtcolorbox{pcchunk}[2]{
  colback=#1!10, colframe=#1, boxrule=0pt, leftrule=2.5pt, arc=1pt, left=5pt, right=5pt, top=2pt, bottom=2pt,
  before skip=2pt, after skip=2pt, fontupper=\footnotesize,
  title={#2}, attach title to upper, coltitle=black!65, fonttitle=\footnotesize\ttfamily\bfseries,
  after title={\enspace},
}
\newcommand{\pcgap}{\par\centerline{\footnotesize\textcolor{black!50}{$\cdots$}}\par}

\titlespacing*{\paragraph}{0pt}{0.5ex plus 0.3ex minus 0.2ex}{0.8em}

%% file: core/1_intro.tex
\section{Introduction}
\label{sec:intro}

In retrieval-augmented generation (RAG), long documents are split into chunks that are indexed and retrieved independently and then passed to the generative LLM. The quality of the generated answer is thus bounded by the quality of chunk retrieval. A chunk may refer to an entity introduced earlier, inherit its meaning from a section heading, or rely on definitions located elsewhere in the document. A standard dense embedding model encodes each chunk in isolation and loses this context, so a chunk may no longer carry enough information to be matched to a query. On the other hand, encoding the entire document as a single vector cannot faithfully capture all the details of a long document. Contextual embedding models represent each chunk of a document as a vector that is computed with respect to the context of the whole document, rather than from the chunk's own text alone.

One of the most common approaches to contextual embedding is late chunking~\citep{gunther2024latechunking}, in which the document is encoded in a single forward pass and chunk embeddings are obtained by pooling the token representations within each chunk. Prior work~\citep{conteb,eslami-etal-2026-diffusion} has shown that such contextualized chunk representations substantially improve retrieval over independently encoded chunks, particularly on long documents. Training these models requires supervision that identifies which parts of a document are relevant to a given query. Prior work relies on gold-chunk annotations: for each query, an annotator, typically a large language model, identifies the chunk of the document that answers it. The model is then trained with the usual cross-document contrastive loss together with a within-document loss that contrasts the gold chunk against the other chunks of the same document.%

Previous work utilizing gold-chunk supervision assumes that relevance is concentrated in a single chunk. Contrasting the gold chunk against its siblings treats every other chunk as a negative, including chunks that introduce an entity, resolve a reference, or supply a complementary fact the answer depends on, while a document-level loss says that the document is relevant but not where. Each annotation yields a single gold label per query, and neither signal lets the model learn to incorporate context selectively. This supervision is also costly to scale. Every training pair requires a large language model to read the document and select a chunk, which limits the size and diversity of the training data. Moreover, labels are tied to the chunking with which they were produced, so changing or varying the chunking strategy requires reannotating the corpus.

The assumption that a query has a single relevant chunk also underlies the evaluation of contextual embeddings. Existing benchmarks, including ConTEB~\citep{conteb}, label one gold passage per query and reward ranking it first. In RAG, however, the generative model might need more than the chunk that states the answer: complementary information, for example, the sentence that identifies the entity, document, date, or version the answer refers to, or a definition or condition stated elsewhere, to produce a correct answer and to confirm that it applies to what was asked. A benchmark with a single gold passage gives no credit for retrieving this information, and does not test whether a model finds the right document among near-identical ones that differ only in such details.

In this work, we address the shortcomings of both the training and the evaluation of contextual embedding models. Our contributions are as follows.
\begin{enumerate}
  \item We propose a novel approach for training contextual embeddings that replaces gold-chunk labels with relevance scores distilled from a query-aware context compression model. 
  \item We introduce \contextbench{} and \qdchunk{}, the first benchmarks to label both answer chunks and the \emph{evidence} chunks needed to interpret or verify the answer. We publish \qdchunk{} as a public leaderboard to encourage further research.
  \item We release \vtwo{} in 9B\footnote{\url{https://huggingface.co/perplexity-ai/pplx-embed-v2-context-9b}} and 0.6B\footnote{\url{https://huggingface.co/perplexity-ai/pplx-embed-v2-context-0.6b}} sizes on Hugging Face and show that our 9B model achieves state-of-the-art performance across various textual chunk retrieval benchmarks, including ConTEB, \contextbench{}, and \qdchunk{}.
  \item We show that our 9B model exhibits zero-shot visual document retrieval (VDR) capabilities and demonstrate that performance scales with chunking granularity. Our models achieve state-of-the-art results among dense retrievers on ViDoRe-V3 and MIRACL-Vision.
\end{enumerate}

%% file: core/3_method.tex
\section{Chunk-level relevance distillation for contextual embeddings}
\label{sec:method}

We address the shortcoming of gold-chunk supervision by distilling chunk-level relevance from a query-aware context compression model. Both of our models, at 9B and 0.6B parameters, are initialized from an in-house vision-language ColBERT model~\citep{v2late}. In this section, we first describe the compression model, which acts as the teacher, and then the training objective of the contextual embedding models.

\subsection{Background: Query-aware context compression}
\label{sec:teacher}

Query-aware context compression selects the spans of a retrieved document that are relevant to the query and passes only those to a generative model~\citep{chirkova2025provence}. As the teacher, we use the context compression model deployed in Perplexity's answer pipeline~\citep{perplexity2026contextcompression}, which formulates compression as token-level sequence labeling. Given a query $q$ and a document $d$, a cross-encoder built on a diffusion-pretrained bidirectional backbone~\citep{eslami-etal-2026-diffusion} encodes the concatenated pair and estimates a relevance score $\rho(v) \in [0, 1]$ for every document token $v \in d$, the probability that the token is kept. Supervision is derived from sentence-level relevance judgments produced by a large language model, which assigns each sentence $s$ of $d$ to one of three classes: \emph{vital}, if $s$ answers $q$ on its own; \emph{relevant}, if $s$ provides context required to interpret the answer; and \emph{irrelevant} otherwise. These classes are mapped to soft targets $y_s = 1$, $y_s \in (0, 1)$ and $y_s = 0$, respectively, and each token $v$ is assigned the target $y_v$ obtained by averaging the targets of the sentences it overlaps, weighted by the extent of the overlap. Following Provence~\citep{chirkova2025provence}, the model is trained to minimize the token-level soft binary cross-entropy over supervised document tokens.

Three properties make this model a suitable teacher. First, $\rho$ is defined per token, so it can be aggregated into a relevance score for any chunking of $d$ at training time. This makes experimentation with different chunking strategies for training feasible. Second, the resulting scores are graded rather than binary, distinguishing answer-bearing, supporting and irrelevant content. Finally, performing inference on this teacher is substantially cheaper than annotation with a large language model, and hence targets can be computed online for every training batch.

\subsection{Training objective of the 9B model}
\label{sec:objective}

The 9B model is trained with a weighted sum of a chunk-level distillation loss and a document-level contrastive loss, over batches of $B$ query--document pairs $\{(q_i, d_i)\}_{i=1}^{|B|}$, in which $d_i$ is relevant to $q_i$ and the other documents in the batch serve as negatives. Each document is chunked with a randomly sampled strategy and encoded in a single forward pass, with separator tokens marking chunk boundaries. We denote by $\mathbf{q}$ the $\ell_2$-normalized embedding of a query $q$, and view a document $d$ as an ordered list of chunks $c$ with $\ell_2$-normalized embeddings $\mathbf{c}$.

The distillation loss teaches the model  within-document relevance.
During training, we pass every positive query--document pair to our context compression model, which assigns relevance scores to every token in the document. We define the relevance of a chunk as the mean score of its $k$ highest-scoring tokens.
For every query within the batch, we obtain a target distribution over the chunks of its positive document by taking the softmax over the target chunk similarities, scaled by a temperature parameter. We extend this target distribution to the chunks of in-batch negative documents by assigning probability zero to them. 
We then train our embedding model via the forward Kullback-Leibler divergence to fit the chunk-level Boltzmann distribution produced by our embedding model to the target. Formally, given $\operatorname{Top}_k(c)$ as the $k$ highest-scoring tokens overlapping chunk $c$, we define:
\begin{gather}
  r(c) := \frac{1}{k}\sum_{v \in \operatorname{Top}_k(c)} \rho(v),
  \qquad
  p^{\mathrm{T}}_i(c) :=
  \begin{cases}
    \dfrac{\exp\big(r(c)/\tau^{\mathrm{T}}\big)}{\sum_{c' \in d_i} \exp\big(r(c')/\tau^{\mathrm{T}}\big)} & c \in d_i,\\[2ex]
    0 & \text{otherwise},
  \end{cases}
  \nonumber\\
  p^{\mathrm{S}}_i(c) := \frac{\exp\big(\mathbf{q}_i^{\top}\mathbf{c}/\tau^{\mathrm{S}}\big)}{\sum_{n=1}^{|B|}\sum_{c' \in d_n} \exp\big(\mathbf{q}_i^{\top}\mathbf{c}'/\tau^{\mathrm{S}}\big)},
  \qquad
  \mathcal{L}_{\mathrm{dist}} := \frac{1}{|B|}\sum_{i=1}^{|B|} \mathrm{KL}\big(p^{\mathrm{T}}_i \,\big\|\, p^{\mathrm{S}}_i\big),
  \label{eq:distill}
\end{gather}
where $r(c), p^{\mathrm{T}}_i(c)$ and $p^{\mathrm{S}}_i(c)$ represent the chunk relevance, teacher and student distributions, respectively. Each token in $\operatorname{Top}_k(c)$ is weighted by its character overlap with $c$, and the student distribution ranges over all chunks of all documents in the batch. Because the teacher places no mass outside $d_i$, the loss decomposes into a within-document term and a document-level term,
\begin{equation}
  \mathrm{KL}\big(p^{\mathrm{T}}_i \,\|\, p^{\mathrm{S}}_i\big)
  = \mathrm{KL}\big(p^{\mathrm{T}}_i \,\|\, \tilde{p}^{\mathrm{S}}_i\big) - \log \sum_{c \in d_i} p^{\mathrm{S}}_i(c),
  \label{eq:kl-decomposition}
\end{equation}
where $\tilde{p}^{\mathrm{S}}_i$ is the student distribution restricted to the chunks of $d_i$ and renormalized. The first term matches the distribution of relevance within the positive document; the second pushes the student's probability mass onto that document and away from competing documents. 
Unlike a single gold-chunk label, our target is graded: supporting chunks within the positive document receive the score the compression model assigns them rather than being treated as negatives, with the intuition that supporting evidence for a query should rank highly, not only the chunk that answers it.

\begin{figure}[t]
  \centering
  \setlength{\parskip}{0pt}
  \begin{minipage}{0.9\textwidth}
    \centering
    \input{figures/training}
  \end{minipage}\par
  \captionsetup{skip=-4pt}
  \caption{%
  The training approach for the 9B model. The contextual encoder (left) embeds each chunk with respect to its document. Query A, a known match for Document A, is scored against the chunks of all documents, and the chunk scores are max-pooled into document scores (right). Chunk scores are trained toward the context compression model's predictions within Document A and zero elsewhere, and document scores toward a one-hot target on Document A. Darker blue indicates higher similarity or relevance.
}
  \label{fig:architecture}
\end{figure}
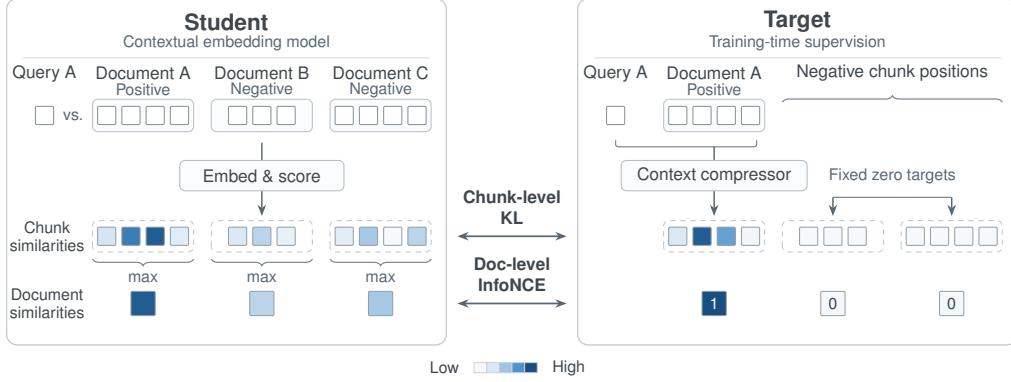

While our embedding model only produces chunk-level similarities, we can naturally derive document-level similarities from them, by considering a document $d$ to be as relevant to a query $q$ as its most relevant chunk:
\begin{equation} s(q, d) := \max_{c \in d} \mathbf{q}^\top \mathbf{c}. \end{equation}
This scoring rule is inspired by the MaxSim operation of ColBERT~\citep{khattab2020colbert}, applied here to chunks instead of tokens. Given these document-level similarities, the contrastive loss is a document-level InfoNCE objective over in-batch negatives,
\begin{equation}
  \mathcal{L}_{\mathrm{con}} := -\frac{1}{|B|}\sum_{i=1}^{|B|} \log \frac{\exp\big(s(q_i, d_i)/\tau\big)}{\sum_{n=1}^{|B|} \exp\big(s(q_i, d_n)/\tau\big)},
  \label{eq:contrastive}
\end{equation}
where $\tau$ is a temperature. Since the gradient of the maximum reaches only the best-matching chunk, the remaining chunks of $d_i$ receive no negative signal from this term. Our overall training objective is $\mathcal{L} := w_{\mathrm{dist}}\,\mathcal{L}_{\mathrm{dist}} + w_{\mathrm{con}}\,\mathcal{L}_{\mathrm{con}}$, with $w_{\mathrm{dist}} = 1$ and $w_{\mathrm{con}} = 2$. Figure~\ref{fig:architecture} shows a schematic of our training approach. 

We employ Matryoshka representation learning (MRL)~\citep{kusupati2022matryoshka} at 1024 and 2048 dimensions, computing both losses on the full embedding and on its first 1024 dimensions, together with quantization-aware training for int8 embeddings~\citep{eslami-etal-2026-diffusion}.

The context compressor serves only as a teacher during training. At inference time, the embedding model retains the standard contextual retrieval interface, producing one contextual embedding per chunk without an additional compression or reranking stage. The approach therefore improves how contextual representations are learned without increasing index storage or introducing teacher-model latency at retrieval time. The same model can also treat the whole document as a single chunk, which yields one mean-pooled vector per document. We refer to this as \emph{single-chunk} and to the default as the \emph{multi-chunk} setting.

\subsection{Distillation to the 0.6B scale}
\label{sec:leaf-distillation}
While we train the 9B contextual model via the context-compressor supervision described above, we find that we achieve better results at the 0.6B scale by distilling our 9B contextual model at the representation level. 
Following~\citet{v2late}, we extend LEAF-style~\citep{leaf} distillation to the contextual setting with multiple chunk embeddings per document. Given a training batch of query-document pairs $(q, d) \in B$, let $\mathbf{q}^T$ and $\mathbf{q}^S$ denote teacher and student query embeddings, and for chunk $c \in d$, let $\mathbf{c}^T$ and $\mathbf{c}^S$ denote teacher and student chunk embeddings.
The training objective is formulated as:
\begin{equation}
    \mathcal{L} := \frac{1}{|B|}\sum_{(q, d) \in B} \left\Vert \mathbf{q}^S - \mathbf{q}^T\right\Vert_2 + \frac{1}{|B|}\sum_{(q, d) \in B} \frac{1}{|d|}\sum_{c \in d} \left\Vert \mathbf{c}^S - \mathbf{c}^T\right\Vert_2,
\end{equation}
where we assume all embeddings to be $\ell_2$-normalized.
We note that, analogous to the results of~\citet{v2late}, the 0.6B and 9B models share an embedding space, which means that documents indexed with the 9B model can be queried with the 0.6B model.

\subsection{Visual document retrieval}
\label{sec:image-chunking-method}
Both models take images through the vision tower of their vision-language base model. The 9B model is trained on text only but we find that visual retrieval capabilities transfer from its base model.
We extend chunking to images, so that a page is represented by several embeddings, each covering one region of it. The vision tower encodes a page as a sequence of visual tokens in row-major order, which we partition into up to 16 horizontal bands, each forming one chunk. Similar to text, \code{<|chunk\_sep|>} tokens are inserted between consecutive bands, and their MRoPE coordinates~\citep{wang2024qwen2} place them centered at the boundary between the two bands. Each chunk embedding is the mean of the token embeddings of its band. Figure~\ref{fig:visual-chunking} (Appendix~\ref{app:visual-chunking}) illustrates partitioning and separator placement, and Section~\ref{sec:vdr-results} shows that chunking improves visual document retrieval (VDR).

\subsection{Training setup and datasets}
\label{sec:training-details}

The 9B model produces 2048-dimensional int8 embeddings, and the 0.6B model, which also outputs int8 embeddings, is trained against their 1024-dimensional MRL prefix. Chunk boundaries are marked with a learned \code{<|chunk\_sep|>} token, and chunk and query embeddings are averages of their token embeddings. Both models are fine-tuned with LoRA~\citep{hu2022lora}. The released models are model soups of several checkpoints from the same run, trained with a warmup--stable--decay learning-rate schedule. Each soup is a uniform average of the model weights. More details of the hyperparameter choices are provided in~Table~\ref{tab:hyperparams}.

The 9B model performs strongly on VDR tasks without training on visual data and we were unable to improve over this performance by including image pairs in its contrastive training. However, we find that the representation-level distillation to 0.6B benefits from visual data and we mix in 25\% image data during its training. 
The text mixture is shared by the 9B and 0.6B models and comprises about 440 public and in-house query--document datasets in over 50 languages, roughly 57M training pairs for the 9B model, combining public retrieval sets, PII-filtered production search logs, and synthetic data in which an LLM generates queries over long documents. For the 9B model, each batch comes from a single dataset to make in-batch negatives harder, while the 0.6B distillation mixes datasets. Context-compression supervision is disabled for datasets without a meaningful notion of within-document relevance, such as FAQ-style data. No training data carries chunk-level annotations, no ConTEB training data is used, and no \contextbench{} or \qdchunk{} data was used for training.

%% file: figures/training.tex
\begingroup
\definecolor{ink}{HTML}{3F464D}
\definecolor{secondary}{HTML}{596673}
\definecolor{muted}{HTML}{B8BEC5}
\definecolor{low}{HTML}{F5F8FC}
\definecolor{lightblue}{HTML}{D9E8F5}
\definecolor{midblue}{HTML}{9FC5E3}
\definecolor{blue}{HTML}{4F93CF}
\definecolor{high}{HTML}{174F82}

\pgfmathsetlengthmacro{\trainingUnit}{\linewidth/20.2}
\pgfmathsetlengthmacro{\trainingBoxWidth}{3.25*\trainingUnit}
\pgfmathsetlengthmacro{\trainingBoxHeight}{.6*\trainingUnit}

\begin{tikzpicture}[
  x=\trainingUnit,y=\trainingUnit,text=ink,inner sep=1.5pt,
  font=\fontfamily{phv}\fontsize{6.5}{8}\selectfont,
  heading/.style={font=\fontfamily{phv}\bfseries\fontsize{8.5}{10}\selectfont},
  label/.style={align=center,font=\fontfamily{phv}\fontsize{6.5}{8}\selectfont},
  small/.style={align=center,font=\fontfamily{phv}\fontsize{6}{7}\selectfont},
  role/.style={small,text=secondary},
  flow/.style={-{Stealth[length=1.5mm]},draw=secondary,line width=.55pt},
  loss/.style={{Stealth[length=1.5mm]}-{Stealth[length=1.5mm]},draw=secondary,line width=.65pt},
  panel/.style={draw=muted!80,fill=white,rounded corners=4pt,line width=.5pt},
  document/.style={draw=muted!85,fill=low!45,rounded corners=2pt,line width=.45pt},
  score group/.style={draw=muted!85,dash pattern=on 2pt off 1.8pt,
    rounded corners=2pt,line width=.4pt},
  chunk/.style={draw=secondary!70,fill=white,line width=.4pt,rounded corners=.5pt},
  relevance/.style={draw=secondary!65,line width=.35pt,rounded corners=.5pt},
  operation/.style={draw=muted,fill=low!50,rounded corners=2pt,
    minimum height=\trainingBoxHeight,minimum width=\trainingBoxWidth,
    inner xsep=6pt,inner ysep=1.5pt,label},
  pooling/.style={decorate,decoration={brace,mirror,amplitude=2.5pt},
    draw=secondary!80,line width=.45pt}
]
\path[use as bounding box] (-.1,-.45) rectangle (20.1,7.25);

\foreach \shift/\title/\subtitle in {
  0/Student/{Contextual embedding model},
  11.3/Target/{Training-time supervision}} {
  \begin{scope}[shift={(\shift,0)}]
    \draw[panel] (0,.3) rectangle (8.7,7.15);
    \node[heading] at (4.35,6.7) {\title};
    \node[role] at (4.35,6.28) {\subtitle};
    \draw[muted!55,line width=.4pt] (.32,5.98) -- (8.38,5.98);
  \end{scope}
}

\node[label] at (.75,5.68) {Query A};
\draw[chunk] (.57,4.66) rectangle ++(.36,.36);
\node[role] at (1.32,4.84) {vs.};
\foreach \left/\center/\letter/\role in {
  1.7/2.7/A/Positive,4.05/5.05/B/Negative,6.4/7.4/C/Negative} {
  \node[label] at (\center,5.68) {Document \letter};
  \node[role] at (\center,5.35) {\role};
  \draw[document] (\left,4.52) rectangle ++(2,.64);
}
\foreach \x in {1.8,2.28,2.76,3.24,4.39,4.87,5.35,6.5,6.98,7.46,7.94}
  \draw[chunk] (\x,4.66) rectangle ++(.36,.36);

\node[operation] (student) at (5.05,3.65) {Embed \& score};
\draw[secondary,line width=.55pt] (5.05,4.3) -- (student.north);
\draw[flow] (student.south) -- (5.05,2.85);

\node[small] at (.8,2.45) {Chunk\\similarities};
\node[small] at (.8,1.12) {Document\\similarities};
\foreach \shift in {0,11.3} {
  \begin{scope}[shift={(\shift,0)}]
    \foreach \left in {1.7,4.05,6.4}
      \draw[score group] (\left,2.14) rectangle ++(2,.62);
  \end{scope}
}
\foreach \left/\center in {1.7/2.7,4.05/5.05,6.4/7.4} {
  \draw[pooling] (\left,2.0) -- ++(2,0);
  \node[role] at (\center,1.62) {max};
}
\foreach \x/\color in {
  1.8/midblue!8!lightblue,2.28/high!32!blue,2.76/high!80!blue,3.24/lightblue!76!low,
  4.39/lightblue!96!low,4.87/midblue!56!lightblue,5.35/lightblue!48!low,
  6.5/lightblue!68!low,6.98/midblue!88!lightblue,7.46/low,7.94/midblue!56!lightblue}
  \draw[relevance,fill=\color] (\x,2.27) rectangle ++(.36,.36);
\foreach \x/\color in {2.7/high!80!blue,5.05/midblue!56!lightblue,7.4/midblue!88!lightblue}
  \draw[relevance,fill=\color] ({\x-.24},.88) rectangle ++(.48,.48);

\node[label] at (12.05,5.68) {Query A};
\draw[chunk] (11.87,4.66) rectangle ++(.36,.36);
\node[label] at (14,5.68) {Document A};
\node[role] at (14,5.35) {Positive};
\draw[document] (13,4.52) rectangle ++(2,.64);
\foreach \x in {13.1,13.58,14.06,14.54}
  \draw[chunk] (\x,4.66) rectangle ++(.36,.36);
\draw[secondary!80,line width=.45pt,rounded corners=1pt]
  (12.05,4.35) -- (12.05,4.22) -- (14.84,4.22) -- (14.84,4.35);
\node[operation] (teacher) at (14,3.65) {Context compressor};
\draw[secondary,line width=.55pt] (14,4.22) -- (teacher.north);
\draw[flow] (teacher.south) -- (14,2.85);

\node[label] at (17.525,5.68) {Negative chunk positions};
\draw[decorate,decoration={brace,amplitude=2.5pt},draw=secondary!80,line width=.45pt]
  (15.35,5.1) -- (19.7,5.1);
\node[role] at (17.525,3.65) {Fixed zero targets};
\draw[secondary!80,line width=.45pt] (16.35,3.22) -- (18.7,3.22);
\foreach \x in {16.35,18.7}
  \draw[flow] (\x,3.22) -- (\x,2.85);

\foreach \x/\color in {13.1/lightblue!92!low,13.58/high!80!blue,
  14.06/blue!72!midblue,14.54/lightblue!12!low}
  \draw[relevance,fill=\color] (\x,2.27) rectangle ++(.36,.36);
\foreach \x in {15.69,16.17,16.65,17.8,18.28,18.76,19.24}
  \draw[relevance,fill=low] (\x,2.27) rectangle ++(.36,.36);
\draw[relevance,fill=high] (13.76,.88) rectangle ++(.48,.48);
\node[small,text=white] at (14,1.12) {1};
\foreach \x in {16.35,18.7} {
  \draw[relevance,fill=low] ({\x-.24},.88) rectangle ++(.48,.48);
  \node[small] at (\x,1.12) {0};
}

\draw[loss] (8.92,2.45) -- (11.08,2.45);
\node[label,font=\fontfamily{phv}\bfseries\fontsize{6.5}{8}\selectfont]
  at (10,3.03) {Chunk-level\\KL};
\draw[loss] (8.92,1.12) -- (11.08,1.12);
\node[label,font=\fontfamily{phv}\bfseries\fontsize{6.5}{8}\selectfont]
  at (10,1.7) {Doc-level\\InfoNCE};

\node[small,anchor=east] at (9.04,-.15) {Low};
\foreach \i/\color in {0/low,1/lightblue,2/midblue,3/blue,4/high}
  \draw[fill=\color,draw=muted!70,line width=.25pt]
    ({9.25+\i*.25},-.25) rectangle ++(.25,.2);
\node[small,anchor=west] at (10.71,-.15) {High};
\end{tikzpicture}
\endgroup

%% file: core/4_benchmark.tex
\section{Evaluating contextual embeddings beyond the gold chunk}
\label{sec:benchmarks}

Existing contextual-retrieval benchmarks, including ConTEB~\citep{conteb}, assume a single relevant passage per query and reward only ranking it first. They do not test whether a model also retrieves the context needed to confirm the answer, or finds the right document among near-identical ones. To fill this gap, we introduce two complementary benchmarks. First, \emph{\contextbench{}} is a controlled benchmark, which is created by adversarially injecting content into long documents to measure each contextual retrieval capability in isolation, with one answer per query. Secondly, \emph{\qdchunk{}} is a benchmark based on real-world use cases, built from LLM-reformulated search queries and labeled by a mixture of LLM judges, with multiple gold passages per query that may span several pages.

\begin{table}[!t]
  \centering
  \small
  \caption{Statistics of \contextbench{} and \qdchunk{}.}
  \label{tab:benchmark-stats}
  \label{tab:pplx-conteb-stats}
  \begin{tabular}{lll}
    \toprule
    & \Contextbench{} & \qdchunk{} \\
    \midrule
    Queries (base / with rewordings) & 2,099 & 6,069 / 11,719 \\
    \quad paraphrased / indirect / scenario & -- & 1,876 / 1,888 / 1,886 \\
    Documents & 38,894 & 71,100 \\
    Chunks & 2.46M sentences & 9.17M of $\leq$128 tokens \\
    Median target document length & $\approx$6,100 tokens & $\approx$4,700 tokens \\
    Languages / domains & -- / 22 & 10 / 26 \\
    Answers per query & 1 & 14.0 spans (mean) \\
    Evidence groups (1 / 2 / $\geq$3 / none) & 964 / 735 / 332 / 68 & 2,799 / 1,577 / 296 / 1,397 \\
    Queries with evidence & 2,031 (97\%) & 4,672 (77\%) \\
    Queries with cross-document gold & -- & 1,300 / 1,102 \\
    \bottomrule
  \end{tabular}
\end{table}

\subsection{\Contextbench{}: a controlled diagnostic via synthetic injection}
\label{sec:benchmark}

\Contextbench{} tests three capabilities of contextual embeddings over long documents. \emph{Document disambiguation} asks whether a model identifies the correct document among near-identical alternatives. \emph{Answer retrieval} asks whether it retrieves answer chunks whose meaning depends on details elsewhere in the document. \emph{Evidence retrieval} asks whether it retrieves the chunks that clarify or substantiate the answer. The queries, documents and capabilities are inspired by conversations with turbopuffer's customers. A property management firm, for example, holds thousands of lease agreements that differ only in tenant names, dates and a few other details. A query such as ``what is the current monthly rent for 5 Park Avenue?'' must be answered from a sentence like ``Monthly rent is \$X,XXX.'', which appears in many documents, while the address that identifies the right lease may be far away in the same document. A good contextual embedding model should use context from across the document to identify the correct chunk and to retrieve both the answer and the sentences that support it. A user or search agent can then verify the answer from a few retrieved sentences instead of the whole document.

Table~\ref{tab:benchmark-stats} summarizes the benchmark. Its domains include corporate filings, clinical research, travel, entertainment, education, government, legal contracts and software documentation. Documents are chunked into sentences with the same boundaries for all models. Each query tests one of twelve types of context dependency, which are listed with examples in Appendix~\ref{app:context-bench-capabilities}. Target documents are long, and 1,061 of the 1,197 distinct target documents have at least 200 sentences. Each query has exactly one correct document; where that document states the answer more than once, as in a summary sentence and a table row, any of these sentences counts. For each query, answer and evidence sentences are labeled in the target document. Long documents often state the same fact in several places, so each evidence group collects the sentences that state the same supporting fact, and retrieving every sentence in any one alternative recovers the group. Evidence groups follow a \emph{reader sufficiency} test. An answer route is a valid combination of an answer chunk and the evidence groups needed to disambiguate or verify it. A query can have several routes, each with its own answer chunks and evidence groups. A model that retrieves the answer sentence of a route and one complete alternative of each of that route’s evidence groups, gives a reader or search agent enough context to confirm the answer without reading the whole document. Queries that need no evidence are excluded from the evidence metrics. \Contextbench{} is held privately by turbopuffer, and all evaluations of the Perplexity embedding models are run by turbopuffer in a blind manner during the development. The benchmark is kept private to mitigate contamination and overfitting.

\subsection{\qdchunk{}: chunk retrieval on real search traffic}
\label{sec:pplx-conteb}

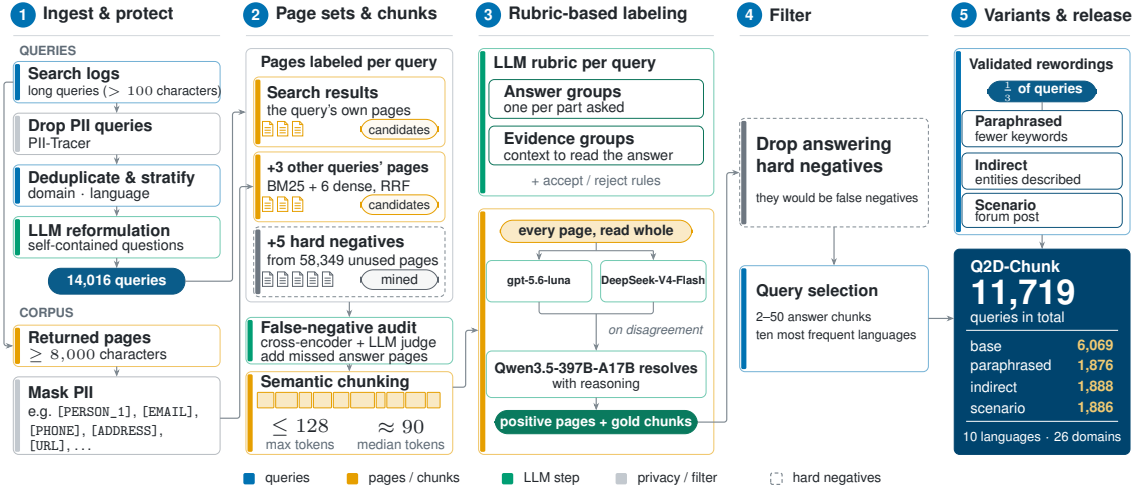
\begin{figure}[t]
  \centering
  \input{figures/pplx_conteb_pipeline_v2}
  \vspace{-6.5mm}
  \caption{The \qdchunk{} construction pipeline, from production search logs to the final benchmark.}
  \label{fig:pplx-conteb-pipeline}
\end{figure}

 In order to measure how well the same capabilities serve information needs that follow the distribution of real user requests, \qdchunk{} complements \contextbench{} with queries from production search traffic. Figure~\ref{fig:pplx-conteb-pipeline} outlines our approach for curating \qdchunk{}.

\paragraph{Queries.} We select long search queries, of more than 100 characters, from production search logs, which our search agent writes from the user's request, to increase the likelihood of including queries whose answers are scattered across a document rather than concentrated in a single chunk. The queries are deduplicated and stratified by domain and language. Like Q2D-Web~\citep{schall2026q2dweb}, we start from the search queries that our search agent writes from users' requests, which are keyword-style; unlike Q2D-Web, an LLM then rewrites each into a self-contained, QA-style question. We call these the \emph{base} queries. Queries containing personally identifiable information (PII), detected with PII-Tracer~\citep{zhang2026piitrace}, are dropped before any query text reaches a language model. This yields 14,016 queries for labeling. Since long search queries tend to repeat the wording of the page they look for, the benchmark also ships three reworded variants of a third of the base queries, each checked by a validator for the same need and no stated answer: \emph{paraphrased}, with fewer shared keywords; \emph{indirect}, with named entities replaced by descriptions; and \emph{scenario}, a first-person forum post. A comparison of these query variants is provided in Figure~\ref{fig:q2d-chunk-variants}.
\label{sec:pplx-conteb-variants}

\paragraph{Corpus and chunking.} The corpus consists of the pages those searches returned that have at least 8,000 characters, with PII masked by typed placeholders such as \code{[PERSON\_1]}, \code{[EMAIL]}, \code{[PHONE]}, \code{[ADDRESS]} or \code{[URL]}. For each query, we label a set of candidate pages. These are its own search results and, since all queries share one corpus, the three pages of other queries most likely to answer it. To find these pages, we rank the corpus with BM25 and six multilingual dense embedders not evaluated here: BGE-M3 \citep{chen2024bgem3}, multilingual-E5-large-instruct \citep{wang2024multilinguale5textembeddings}, gte-multilingual-base \citep{zhang-etal-2024-mgte}, Snowflake Arctic-Embed-L~v2.0 \citep{yu2025arcticembed}, and Qwen3-Embedding-4B and -8B \citep{qwen3-embedding}. We fuse the rankings by reciprocal rank fusion. A candidate page becomes a positive page when labeling finds it to be about the query's subject and to state an answer; answer-like text on pages about another subject becomes a distractor. Labeling all candidates, rather than only the pages a search returned, reduces false negatives. To make the corpus more challenging, we also mine five hard negatives per query with the same seven retrievers and fusion, applied to 58,349 fetched pages that belong to no query and duplicate no corpus page. Finally, to catch false negatives that remain, including on pages mined as hard negatives for other queries, a cross-encoder (gte-multilingual-reranker-base) scores, for each query, its next 20 corpus pages and 10 hard-negative pages under the fused rankings, and every page in an evaluated model's top 10; an LLM judge (Qwen3.5-397B-A17B) reviews the pages above a threshold that keeps 95\% of known answer pages, and the flagged pages are labeled like the candidates, adding answer pages to 1,175 queries. Pages are split by a semantic text splitter into chunks of at most 128 tokens that end at the strongest available boundary, e.g., paragraph, line, sentence, and word.

\begin{figure}[!t]
  \centering
  \includegraphics[width=\textwidth]{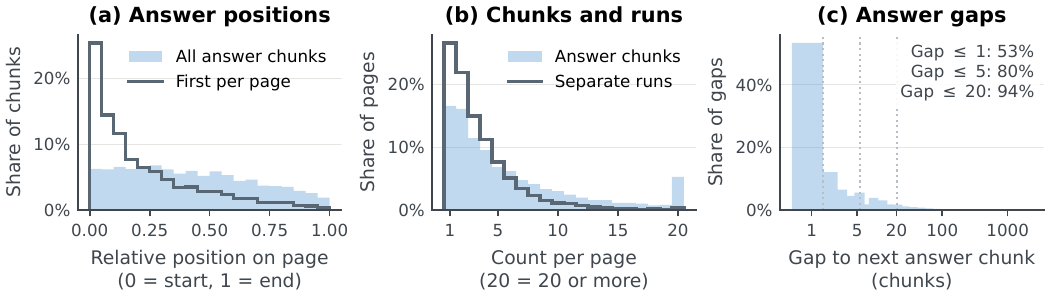}
  \caption{Where the answer chunks of \qdchunk{} sit within their pages (128-token chunks, one point per query and page with answers). (a) Relative position on the page of all answer chunks and of the first answer chunk of each page. (b) Answer chunks per page and the number of separate runs of adjacent answer chunks they form. (c) Gap between consecutive answer chunks of a page.}
  \label{fig:pplx-conteb-gold-positions}
\end{figure}

\paragraph{Labels.} For each query, an LLM first writes a rubric of answer groups, one per part the query asks for, and evidence groups, one per piece of context a reader needs to interpret or verify the answer: the entity, document or period it belongs to (identification), a qualifier that limits where it holds (condition), or a definition or unit needed to read it (interpretation), each with accept and reject rules. Evidence is labeled only on pages that hold answers. Every chunk of every candidate page is labeled against this rubric, which decides which candidates are positive pages and marks the gold chunks within them. Mined hard negatives are labeled in the same way, and those that answer the query are removed as false negatives. The labelers see each page whole, so that no label misses the context it depends on. Every page is labeled by two independent labelers (gpt-5.6-luna and DeepSeek-V4-Flash), and wherever they disagree, Qwen3.5-397B-A17B with reasoning resolves the conflict. We keep queries with 2 to 50 answer chunks in the ten most frequent languages. Figure~\ref{fig:pplx-conteb-example} shows a labeled example.

\paragraph{Statistics.} The final benchmark has 6,069 base queries, 11,719 with the reworded variants, in 10 languages and 26 top-level domains (Table~\ref{tab:pplx-conteb-stats}); 71\% are English (Figure~\ref{fig:q2d-chunk-languages}), and the largest domains are health, law and finance (Figure~\ref{fig:q2d-chunk-domains}). 3,660 queries have answers on at least two pages, and within a page, a median of four answer chunks form three runs spanning 45\% of it (Figure~\ref{fig:pplx-conteb-gold-positions}). We publish \qdchunk{} as a public leaderboard\footnote{\url{https://huggingface.co/spaces/perplexity-ai/q2d-chunk-leaderboard}}.

\subsection{Metrics}
\label{sec:pplx-conteb-metrics}

The metrics are grouped by the capability they measure. Recall metrics count each answer or evidence group once, so a group is recovered when any one of its alternatives is retrieved.

\paragraph{Answer retrieval.} For each query, \textbf{Answer@$K$} is the fraction of its answer groups with at least one accepted answer chunk among the top $K$ chunks of the corpus. In \contextbench{}, a query has a single answer group, made of its accepted answer sentences. In \qdchunk{}, each part of the question forms its own answer group. Answer@10 is the main score of \qdchunk{}, whose answers span many chunks and pages. It is complete only when every answer group is found. For comparability with other benchmarks, we also report \textbf{nDCG@10} over the answer chunks. We choose Answer@10 as our main score since nDCG@10 credits every answer chunk, so a ranking can score well on it by drawing many chunks from a single page, even when parts of the question remain unanswered.

\paragraph{Evidence retrieval.} For queries that have evidence, \textbf{Evidence recall@$K$} is the fraction of their evidence groups recovered among the top $K$ chunks. Accepted answer chunks are first removed from the ranking. This way, answers do not occupy the slots on which evidence is measured. \textbf{All-Evidence@$K$} is the fraction of these queries with every evidence group recovered. We compute evidence recall in two ways. The corpus variant ranks all chunks of the corpus and measures whether evidence reaches a reader from the whole corpus. The in-document variant mirrors an agent that reads the top chunks of a document it has retrieved. It ranks the chunks within each gold document that the model places among its top 10 documents, and measures whether a model surfaces evidence once the right document is found. \qdchunk{} reports both variants. \Contextbench{} reports the in-document variant only, because its evidence sentences are often generic and would compete with near-identical sentences of look-alike documents. Answer@$K$ needs no such condition, since an answer among the top $K$ already places its document there.

\paragraph{Coverage and reading cost.} On \qdchunk{}, \textbf{$\alpha$-nDCG@10}~\citep{clarke2008novelty} measures how many different aspects of a query a ranking covers. It uses the answer and evidence groups as aspects, with answer groups weighted twice as much as evidence groups. Each chunk that repeats an already covered group is discounted by a factor $1-\alpha$, with $\alpha = 0.5$. \textbf{Tokens to full coverage} is the number of tokens a reader goes through, reading the ranked chunks from the top, until every answer and evidence group is covered. We report its median over queries, and a query whose groups are never all covered counts as unreached. This metric complements the recalls. Answer@10 and Evidence@10 measure partial coverage at a fixed depth of ten chunks. Tokens to full coverage instead measures the depth a ranking needs for complete coverage. It is therefore set by each query's hardest group, which a top-10 recall does not see. It is also counted in tokens, the budget that matters when retrieved chunks fill a language model's context.

\paragraph{Document retrieval.} \textbf{Document@$K$} scores each document by its highest-ranked chunk and measures whether the gold document is among the top $K$ documents. On \contextbench{}, whose corpus contains look-alike documents that share the gold document's vocabulary, this tests document disambiguation: whether context lets the gold document beat near-identical alternatives. On \qdchunk{}, whose answers span several pages, it is the fraction of a query's answer pages among the top $K$ pages. Non-contextual models embed every sentence (\contextbench{}) or chunk (\qdchunk{}) on its own. As a single-vector baseline, we also embed each whole document, truncated to 32,768 tokens, as one vector with our models.

%% file: figures/pplx_conteb_pipeline_v2.tex
\begingroup
\definecolor{ink}{HTML}{22272C}
\definecolor{secondary}{HTML}{5F6B76}
\definecolor{muted}{HTML}{C3C9CF}
\definecolor{blue}{HTML}{0072B2}
\definecolor{teal}{HTML}{009E73}
\definecolor{gold}{HTML}{E69F00}
\definecolor{lavender}{HTML}{6E7781}
\definecolor{darkblue}{HTML}{00446E}

\newcommand{\pageicon}[3]{%
  \begin{scope}[shift={(#1,#2)}]
    \draw[draw=#3,fill=white,line width=0.27pt]
      (0,0)--(.2,0)--(.2,.22)--(.14,.28)--(0,.28)--cycle;
    \draw[draw=#3,line width=0.24pt] (.14,.28)--(.14,.22)--(.2,.22);
    \foreach \yy in {.06,.11,.16}
      \draw[draw=#3,line width=0.2pt] (.04,\yy)--(.16,\yy);
  \end{scope}%
}

\pgfmathsetlengthmacro{\pipelineUnit}{\linewidth/21.94}
\begin{tikzpicture}[
  x=\pipelineUnit,y=\pipelineUnit,text=ink,inner sep=1.5pt,
  font=\fontfamily{phv}\fontsize{5.78}{6.8}\selectfont,
  title/.style={anchor=west,font=\fontfamily{phv}\bfseries\fontsize{6.12}{7.48}\selectfont},
  body/.style={anchor=west,font=\fontfamily{phv}\fontsize{5.44}{6.8}\selectfont},
  small/.style={font=\fontfamily{phv}\fontsize{5.1}{6.12}\selectfont},
  card/.style={rounded corners=2.04pt,line width=0.37pt},
  query/.style={card,draw=blue!55!white,fill=white},
  pages/.style={card,draw=gold!60!white,fill=white},
  llm/.style={card,draw=teal!55!white,fill=white},
  privacy/.style={card,draw=secondary!45,fill=white},
  negative/.style={card,draw=secondary!80,fill=white,dash pattern=on 1.4pt off 1pt},
  flow/.style={-{Stealth[length=1.02mm]},draw=secondary!80,line width=0.44pt,rounded corners=1.02pt},
  pipe/.style={draw=secondary!80,line width=0.41pt,rounded corners=1.02pt},
  pill/.style={rounded corners=4.76pt,line width=0.34pt},
  stage/.style={anchor=west,font=\fontfamily{phv}\bfseries\fontsize{6.46}{7.48}\selectfont}
]
\path[use as bounding box] (-.26,-.3) rectangle (21.68,9.55);

\foreach \xx/\ww/\nn/\tt in {
  0/4/1/{Ingest \& protect},
  4.5/4/2/{Page sets \& chunks},
  9/4.55/3/{Rubric-based labeling},
  14.05/3.65/4/{Filter},
  18.2/3.4/5/{Variants \& release}} {
  \node[circle,fill=blue,minimum size=9.52pt,inner sep=0pt,text=white,
    font=\fontfamily{phv}\bfseries\fontsize{5.78}{6.8}\selectfont]
    at ({\xx+.19},9.1) {\nn};
  \node[stage] at ({\xx+.49},9.1) {\tt};
  \draw[muted!75,line width=0.34pt] (\xx,8.7)--++(\ww,0);
}

\node[body,text=secondary,font=\fontfamily{phv}\bfseries\fontsize{4.76}{5.44}\selectfont]
  at (.03,8.4) {QUERIES};
\draw[query] (0,7.4) rectangle (4,8.2);
\fill[blue,rounded corners=0.6pt] (0.030,7.450) rectangle (0.130,8.150);
\node[title] at (.2,7.94) {Search logs};
\node[body,font=\fontfamily{phv}\fontsize{4.9}{6.12}\selectfont]
  at (.2,7.62) {long queries ($>100$ characters)};
\draw[privacy] (0,6.4) rectangle (4,7.2);
\fill[muted,rounded corners=0.6pt] (0.030,6.450) rectangle (0.130,7.150);
\node[title] at (.2,6.94) {Drop PII queries};
\node[body] at (.2,6.62) {PII-Tracer};
\draw[query] (0,5.4) rectangle (4,6.2);
\fill[blue,rounded corners=0.6pt] (0.030,5.450) rectangle (0.130,6.150);
\node[title] at (.2,5.94) {Deduplicate \& stratify};
\node[body] at (.2,5.62) {domain $\cdot$ language};
\draw[llm] (0,4.4) rectangle (4,5.2);
\fill[teal,rounded corners=0.6pt] (0.030,4.450) rectangle (0.130,5.150);
\node[title,text=ink] at (.2,4.94) {LLM reformulation};
\node[body] at (.2,4.62) {self-contained questions};
\foreach \yy in {7.4,6.4,5.4}
  \draw[flow] (2,\yy)--++(0,-.2);
\draw[flow] (2,4.4)--(2,4.2);
\draw[pill,draw=blue!65!ink,fill=blue!70!ink] (.68,3.7) rectangle (3.32,4.2);
\node[text=white,font=\fontfamily{phv}\bfseries\fontsize{5.44}{6.8}\selectfont]
  at (2,3.95) {14,016 queries};

\node[body,text=secondary,font=\fontfamily{phv}\bfseries\fontsize{4.76}{5.44}\selectfont]
  at (.03,3.34) {CORPUS};
\draw[pages] (0,2.3) rectangle (4,3.1);
\fill[gold,rounded corners=0.6pt] (0.030,2.350) rectangle (0.130,3.050);
\node[title,text=ink] at (.2,2.84) {Returned pages};
\node[body] at (.2,2.52) {$\geq8{,}000$ characters};
\draw[flow] (2,2.3)--(2,2.1);
\draw[privacy] (0,.6) rectangle (4,2.1);
\fill[muted,rounded corners=0.6pt] (0.030,0.650) rectangle (0.130,2.050);
\node[title] at (.2,1.8) {Mask PII};
\node[body] at (.2,1.4) {e.g.\ \texttt{[PERSON\_1]}, \texttt{[EMAIL]},};
\node[body] at (.2,1.06) {\texttt{[PHONE]}, \texttt{[ADDRESS]},};
\node[body] at (.2,.76) {\texttt{[URL]}, \ldots};
\draw[flow] (0,7.8)--(-.18,7.8)--(-.18,2.7)--(0,2.7);

\draw[card,draw=muted,fill=white] (4.5,3.55) rectangle (8.5,8.45);
\node[title,font=\fontfamily{phv}\bfseries\fontsize{5.78}{6.8}\selectfont]
  at (4.7,8.16) {Pages labeled per query};
\draw[pages] (4.65,6.55) rectangle (8.35,7.9);
\fill[gold,rounded corners=0.6pt] (4.680,6.600) rectangle (4.780,7.850);
\node[title,text=ink] at (4.82,7.6) {Search results};
\node[body] at (4.82,7.23) {the query's own pages};
\foreach \jj in {0,1,2} {\pageicon{4.82+\jj*.29}{6.76}{gold}}
\draw[pill,draw=gold,fill=gold!5] (6.72,6.7) rectangle (8.22,7.08);
\node[small,text=ink,font=\fontfamily{phv}\fontsize{4.76}{5.44}\selectfont]
  at (7.47,6.89) {candidates};

\draw[pages] (4.65,5.1) rectangle (8.35,6.45);
\fill[gold,rounded corners=0.6pt] (4.680,5.150) rectangle (4.780,6.400);
\node[title,text=ink,font=\fontfamily{phv}\bfseries\fontsize{5.44}{6.8}\selectfont]
  at (4.82,6.14) {+3 other queries' pages};
\node[body] at (4.82,5.78) {BM25 + 6 dense, RRF};
\foreach \jj in {0,1,2} {\pageicon{4.82+\jj*.29}{5.3}{gold}}
\draw[pill,draw=gold,fill=gold!5] (6.72,5.24) rectangle (8.22,5.62);
\node[small,text=ink,font=\fontfamily{phv}\fontsize{4.76}{5.44}\selectfont]
  at (7.47,5.43) {candidates};

\draw[negative] (4.65,3.65) rectangle (8.35,5.0);
\fill[lavender,rounded corners=0.6pt] (4.680,3.700) rectangle (4.780,4.950);
\node[title,text=ink] at (4.82,4.7) {+5 hard negatives};
\node[body] at (4.82,4.33) {from 58,349 unused pages};
\foreach \jj in {0,...,4} {\pageicon{4.82+\jj*.29}{3.85}{lavender!75!ink}}
\draw[pill,draw=lavender!75!ink,fill=lavender!6] (6.72,3.79) rectangle (8.22,4.17);
\node[small,text=ink] at (7.47,3.98) {mined};

\draw[flow] (3.32,3.95)--(4.24,3.95)--(4.24,7.22)--(4.5,7.22);
\draw[flow] (4,1.3)--(4.38,1.3)--(4.38,5.78)--(4.65,5.78);
\draw[flow] (6.5,3.55)--(6.5,3.25);
\draw[llm] (4.5,2.35) rectangle (8.5,3.25);
\fill[teal,rounded corners=0.6pt] (4.530,2.400) rectangle (4.630,3.200);
\node[title,text=ink] at (4.7,3.02) {False-negative audit};
\node[body] at (4.7,2.74) {cross-encoder + LLM judge};
\node[body] at (4.7,2.48) {add missed answer pages};
\draw[flow] (6.5,2.35)--(6.5,2.2);
\draw[pages] (4.5,.6) rectangle (8.5,2.2);
\fill[gold,rounded corners=0.6pt] (4.530,0.650) rectangle (4.630,2.150);
\node[title,text=ink] at (4.7,1.98) {Semantic chunking};
\foreach \xx/\ww in {4.72/.32,5.075/.42,5.53/.24,5.805/.35,6.19/.30,
  6.525/.46,7.02/.17,7.225/.33,7.59/.38,8.005/.25}
  \draw[fill=gold!23,draw=gold,line width=0.24pt,rounded corners=0.34pt]
    (\xx,1.5) rectangle ++(\ww,.3);
\node[text=ink,font=\fontfamily{phv}\bfseries\fontsize{7.48}{8.84}\selectfont]
  at (5.55,1.13) {$\leq128$};
\node[text=ink,font=\fontfamily{phv}\bfseries\fontsize{7.48}{8.84}\selectfont]
  at (7.5,1.13) {$\approx90$};
\node[small,text=secondary] at (5.55,.8) {max tokens};
\node[small,text=secondary] at (7.5,.8) {median tokens};

\draw[llm] (9,5.6) rectangle (13.55,8.45);
\fill[teal,rounded corners=0.6pt] (9.030,5.650) rectangle (9.130,8.400);
\node[title,text=ink] at (9.2,8.15) {LLM rubric per query};
\draw[card,draw=teal!65!ink,fill=white] (9.2,7.1) rectangle (13.35,7.85);
\node[title] at (9.4,7.59) {Answer groups};
\node[body] at (9.4,7.29) {one per part asked};
\draw[card,draw=teal!65!ink,fill=white] (9.2,6.2) rectangle (13.35,6.95);
\node[title] at (9.4,6.69) {Evidence groups};
\node[body] at (9.4,6.39) {context to read the answer};
\node[small,text=secondary] at (11.275,5.88) {+ accept / reject rules};

\draw[pages] (9,.6) rectangle (13.55,5.35);
\fill[gold,rounded corners=0.6pt] (9.030,0.650) rectangle (9.130,5.300);
\draw[pill,fill=gold!22,draw=gold] (9.42,4.7) rectangle (13.13,5.13);
\node[text=ink,font=\fontfamily{phv}\bfseries\fontsize{5.3}{6.12}\selectfont]
  at (11.275,4.915) {every page, read whole};
\draw[pipe] (11.275,4.7)--(11.275,4.54);
\draw[pipe] (10.17,4.54)--(12.38,4.54);
\draw[flow] (10.17,4.54)--(10.17,4.35);
\draw[flow] (12.38,4.54)--(12.38,4.35);
\draw[llm] (9.15,3.55) rectangle (11.19,4.35);
\draw[llm] (11.36,3.55) rectangle (13.4,4.35);
\node[text=ink,font=\fontfamily{phv}\bfseries\fontsize{4.22}{5.44}\selectfont]
  at (10.17,3.95) {gpt-5.6-luna};
\node[text=ink,font=\fontfamily{phv}\bfseries\fontsize{4.22}{5.44}\selectfont]
  at (12.38,3.95) {DeepSeek-V4-Flash};
\draw[pipe] (10.17,3.55)--(10.17,3.32)--(12.38,3.32)--(12.38,3.55);
\draw[flow] (11.275,3.32)--(11.275,2.6);
\node[anchor=west,text=secondary,font=\fontfamily{phv}\itshape\fontsize{4.76}{6.12}\selectfont]
  at (11.42,2.98) {on disagreement};
\draw[llm] (9.15,1.7) rectangle (13.4,2.6);
\node[text=ink,font=\fontfamily{phv}\bfseries\fontsize{5.44}{6.8}\selectfont]
  at (11.275,2.27) {Qwen3.5-397B-A17B resolves};
\node[small] at (11.275,1.95) {with reasoning};
\draw[flow] (11.275,1.7)--(11.275,1.45);
\draw[pill,draw=teal!70!ink,fill=teal!70!ink] (9.32,1.0) rectangle (13.23,1.45);
\node[text=white,font=\fontfamily{phv}\bfseries\fontsize{5.17}{6.12}\selectfont]
  at (11.275,1.225) {positive pages + gold chunks};
\draw[flow] (8.5,1.9)--(8.75,1.9)--(8.75,3.0)--(9,3.0);

\draw[negative] (14.05,5.0) rectangle (17.7,7.1);
\fill[lavender,rounded corners=0.6pt] (14.080,5.050) rectangle (14.180,7.050);
\node[align=left,anchor=west,text=ink,
  font=\fontfamily{phv}\bfseries\fontsize{6.8}{8.16}\selectfont]
  at (14.28,6.36) {Drop answering\\hard negatives};
\node[body,font=\fontfamily{phv}\fontsize{4.69}{6.12}\selectfont]
  at (14.28,5.55) {they would be false negatives};
\draw[flow] (13.23,1.225)--(13.8,1.225)--(13.8,6.05)--(14.05,6.05);
\draw[flow] (15.875,5.0)--(15.875,4.25);
\draw[query] (14.05,2.2) rectangle (17.7,4.25);
\fill[blue,rounded corners=0.6pt] (14.080,2.250) rectangle (14.180,4.200);
\node[title] at (14.28,3.75) {Query selection};
\node[body,font=\fontfamily{phv}\fontsize{4.76}{6.12}\selectfont] at (14.28,3.27) {2--50 answer chunks};
\node[body,font=\fontfamily{phv}\fontsize{4.76}{6.12}\selectfont] at (14.28,2.9) {ten most frequent languages};

\draw[query] (18.2,4.85) rectangle (21.6,8.45);
\fill[blue,rounded corners=0.6pt] (18.230,4.900) rectangle (18.330,8.400);
\node[title,font=\fontfamily{phv}\bfseries\fontsize{5.44}{6.8}\selectfont]
  at (18.4,8.13) {Validated rewordings};
\draw[pill,draw=blue!65!ink,fill=blue!70!ink] (18.85,7.42) rectangle (20.95,7.82);
\node[text=white,font=\fontfamily{phv}\bfseries\fontsize{5.1}{6.12}\selectfont]
  at (19.9,7.62) {$\frac{1}{3}$ of queries};
\draw[pipe] (19.9,7.42)--(19.9,7.25);
\foreach \bottom/\top/\name/\detail in {
  6.55/7.25/Paraphrased/{fewer keywords},
  5.73/6.43/Indirect/{entities described},
  4.99/5.65/Scenario/{forum post}} {
  \draw[card,draw=blue!60!ink,fill=white] (18.35,\bottom) rectangle (21.45,\top);
  \node[title,font=\fontfamily{phv}\bfseries\fontsize{5.44}{6.8}\selectfont]
    at (18.52,{\top-.23}) {\name};
  \node[body,font=\fontfamily{phv}\fontsize{5.1}{6.12}\selectfont]
    at (18.52,{\bottom+.19}) {\detail};
}
\draw[flow] (19.9,4.85)--(19.9,4.57);
\draw[card,draw=darkblue,fill=darkblue] (18.2,.6) rectangle (21.6,4.57);
\node[title,text=white] at (18.43,4.24) {\qdchunk{}};
\node[anchor=west,text=white,font=\fontfamily{phv}\bfseries\fontsize{12.92}{14.28}\selectfont]
  at (18.43,3.69) {11,719};
\node[body,text=white] at (18.43,3.23) {queries in total};
\draw[white!40,line width=0.27pt] (18.43,3.0)--(21.37,3.0);
\foreach \yy/\name/\number in {
  2.7/base/{6,069},2.3/paraphrased/{1,876},1.9/indirect/{1,888},1.5/scenario/{1,886}} {
  \node[body,text=white] at (18.43,\yy) {\name};
  \node[anchor=east,text=gold!55!white,font=\fontfamily{phv}\bfseries\fontsize{5.44}{6.8}\selectfont]
    at (21.37,\yy) {\number};
}
\draw[white!40,line width=0.27pt] (18.43,1.24)--(21.37,1.24);
\node[text=white,font=\fontfamily{phv}\fontsize{4.9}{6.12}\selectfont]
  at (19.9,.92) {10 languages $\cdot$ 26 domains};
\draw[flow] (17.7,3.23)--(18.2,3.23);

\def\negkind{negative}\foreach \xx/\kind/\kindcol/\tt in {
  4.45/query/blue/queries,6.45/pages/gold/{pages / chunks},9.45/llm/teal/{LLM step},
  11.65/privacy/muted/{privacy / filter},14.65/negative/lavender/{hard negatives}} {
  \ifx\kind\negkind\draw[secondary!80,dash pattern=on 1.4pt off 1pt,line width=0.45pt,rounded corners=0.68pt] (\xx,.02) rectangle ++(.22,.22);\else\fill[\kindcol,rounded corners=0.68pt] (\xx,.02) rectangle ++(.22,.22);\fi
  \node[body,font=\fontfamily{phv}\fontsize{5.1}{6.12}\selectfont]
    at ({\xx+.34},.13) {\tt};
}
\end{tikzpicture}
\endgroup

%% file: core/5_results.tex
\section{Results}
\label{sec:results}

We evaluate our two models against contextual and non-contextual embedding models. We first report results on \contextbench{} and \qdchunk{}, which directly test the capabilities that motivate this work, and on ConTEB, to our knowledge the only previously available public benchmark for contextual embeddings. We then evaluate document-level retrieval, retrieval quality as a function of storage per vector, and visual document retrieval. Domain-specific chunk and document retrieval and sensitivity to the chunk size (Appendix~\ref{app:q2d}) and results on BRIGHT (Appendix~\ref{app:bright}) are provided in the appendix. All evaluations encode documents of up to 32,768 tokens in a single pass.

\subsection{Answer and evidence retrieval}
\label{sec:results-benchmarks}
\begin{table}[t]
  \centering
  \small  \caption{Results on \contextbench{}: Answer@$K$, evidence recall@$K$ within retrieved documents, and Document@$K$. All-Evidence@10 is the fraction of queries with every evidence group recovered. Answer chunks are removed before evidence is ranked. $^{\dagger}$Non-contextual baseline, which encodes every sentence independently. Scores are in \%.}
  \label{tab:context-bench-results}
  \resizebox{\textwidth}{!}{%
  \begin{tabular}{lcccccccccc}
    \toprule
    & \multicolumn{3}{c}{Answer@$K$} & \multicolumn{4}{c}{Evidence recall@$K$ (in document)} & \multicolumn{3}{c}{Document@$K$} \\
    \cmidrule(lr){2-4} \cmidrule(lr){5-8} \cmidrule(lr){9-11}
    Model & @3 & @5 & @10 & @3 & @5 & @10 & All-Evidence@10 & @3 & @5 & @10 \\
    \midrule
    \vtwosize{9b} & \textbf{23.6} & \textbf{31.1} & \textbf{45.5} & \textbf{23.2} & \textbf{30.8} & \textbf{40.6} & \textbf{31.1} & 29.7 & 40.3 & \textbf{61.6} \\\
    \vtwosize{0.6b} & 9.3 & 14.0 & 21.9 & 14.4 & 18.8 & 23.8 & 18.6 & 16.4 & 23.7 & 39.0 \\
    \vonesize{4b} & 12.1 & 17.2 & 26.4 & 22.0 & 28.4 & 36.4 & 28.6 & \textbf{29.9} & \textbf{40.5} & 58.1 \\
    voyage-context-4 & 15.2 & 20.2 & 31.1 & 21.6 & 27.1 & 35.6 & 27.6 & 24.2 & 33.4 & 51.6 \\
    voyage-4-large$^{\dagger}$ & 2.7 & 4.1 & 6.4 & 16.7 & 19.7 & 23.4 & 18.8 & 17.6 & 25.4 & 40.2 \\
    \bottomrule
  \end{tabular}}
\end{table}

\paragraph{\Contextbench{}.} Table~\ref{tab:context-bench-results} reports \contextbench{}, on which all our models were evaluated blindly by turbopuffer. \vtwosize{9b} leads answer retrieval at every $K$ and outperforms the strongest baseline, voyage-context-4, by 14.4 points at Answer@10. Non-contextual models rarely rank the answer sentence highly. For evidence within retrieved documents, \vtwosize{9b} is also the strongest model. The 0.6B model trails the larger contextual models on this benchmark. In the single-chunk setting, which embeds each document as one vector, Document@10 is 29.8 points lower for the 9B model and 8.8 points for the 0.6B model than when each document is ranked by its best chunk.

\label{sec:results-pplx-conteb}

\paragraph{\qdchunk{}.} Table~\ref{tab:pplx-conteb-results} reports \qdchunk{} at 128-token chunks. Besides our two models, we evaluate voyage-context-4~\citep{voyagecontext4_2026} and its non-contextual counterpart voyage-4-large~\citep{voyage4_2026}, both served through Voyage's API\footnote{\url{https://docs.voyageai.com/docs/embeddings}}. We also compare our results with the \vone{} model, as well as the non-contextual \texttt{pplx-embed-v1-4b}. When using non-contextual models, every chunk is encoded separately. The training and test data of \vtwo{} are fully disjoint and drawn from different dates, which rules out data leakage.

\begin{table}[t]
  \centering
  \small
  \caption{Answer@10 on the four query configurations of \qdchunk{}, and evidence, combined and page metrics and median tokens to full coverage on the base queries. $^{\dagger}$Non-contextual model, which encodes every chunk independently. Scores are in \%.}
  \label{tab:pplx-conteb-results}
  \resizebox{\textwidth}{!}{%
  \begin{tabular}{lcccccccccc}
    \toprule
    & \multicolumn{4}{c}{Answers (Answer@10)} & \multicolumn{3}{c}{Evidence@10} & \multicolumn{2}{c}{Both} & Pages \\
    \cmidrule(lr){2-5} \cmidrule(lr){6-8} \cmidrule(lr){9-10} \cmidrule(lr){11-11}
    & & & & & & & & & Tokens & \\
    Model & base & paraphr. & indirect & scenario & corpus & in document & All-Evidence & $\alpha$-nDCG@10 & to coverage & Document@10 \\
    \midrule
    \vtwosize{9b} & \textbf{86.4} & \textbf{85.3} & \textbf{82.6} & \textbf{83.9} & 46.6 & 72.6 & 37.5 & \textbf{68.1} & \textbf{1,954} & \textbf{98.4} \\
    \vtwosize{0.6b} & 83.1 & 81.6 & 77.3 & 79.9 & 44.0 & 69.5 & 35.2 & 64.2 & 2,264 & 97.9 \\
    voyage-context-4 & 83.9 & 83.2 & 80.2 & 80.9 & 41.8 & 68.3 & 33.1 & 64.7 & 2,412 & 97.6 \\
    \vonesize{4b} & 74.9 & 73.8 & 65.4 & 70.5 & 41.6 & 71.5 & 33.1 & 53.9 & 2,826 & 95.7 \\
    \vonesize{0.6b} & 68.1 & 65.9 & 54.3 & 61.9 & 38.6 & 65.8 & 30.7 & 47.4 & 3,549 & 91.3 \\
    voyage-4-large$^{\dagger}$ & 74.5 & 73.2 & 69.5 & 72.0 & 39.6 & 72.4 & 30.9 & 53.0 & 3,425 & 94.6 \\
    Nemotron-3-Embed-8B$^{\dagger}$ & 73.6 & 71.7 & 68.1 & 70.7 & \textbf{49.3} & \textbf{78.3} & \textbf{39.8} & 53.2 & 2,577 & 96.1 \\
    \texttt{pplx-embed-v1-4b}$^{\dagger}$ & 68.1 & 67.4 & 61.4 & 64.7 & 46.0 & 76.3 & 36.7 & 47.2 & 3,308 & 95.4 \\
    \bottomrule
  \end{tabular}}
\end{table}

\begin{table}[t]
  \centering
  \small
  \caption{Answer@10 on \qdchunk{} per top-level domain, for the ten most frequent labeled domains (base queries; queries per domain in parentheses). Scores are in \%. $^{\dagger}$Non-contextual model, which encodes every chunk independently.}
  \label{tab:q2d-chunk-domain-results}
  \resizebox{\textwidth}{!}{%
  \begin{tabular}{lcccccccccc}
    \toprule
    & Health & Law & Finance & Business & Technology & Education & Politics & Science & Consumer goods & Real estate \\
    Model & (1,456) & (1,239) & (719) & (465) & (456) & (413) & (190) & (177) & (145) & (119) \\
    \midrule
    \vtwosize{9b} & \textbf{87.7} & \textbf{86.9} & \textbf{85.5} & \textbf{86.1} & \textbf{86.8} & \textbf{86.1} & \textbf{83.9} & \textbf{86.5} & \textbf{82.2} & \textbf{85.9} \\
    \vtwosize{0.6b} & 84.9 & 83.3 & 83.2 & 83.3 & 83.3 & 82.3 & 80.7 & 82.7 & 77.7 & 81.9 \\
    voyage-context-4 & 84.1 & 85.0 & 84.3 & 82.0 & 85.6 & 83.8 & 80.9 & 81.6 & 80.7 & 84.3 \\
    \vonesize{4b} & 73.1 & 76.5 & 77.5 & 75.4 & 76.0 & 71.4 & 77.0 & 72.3 & 69.0 & 74.7 \\
    \vonesize{0.6b} & 68.0 & 67.4 & 70.0 & 71.3 & 69.3 & 62.9 & 72.6 & 66.4 & 56.8 & 75.5 \\
    voyage-4-large$^{\dagger}$ & 74.4 & 75.6 & 73.7 & 73.8 & 77.0 & 68.9 & 74.6 & 79.0 & 71.3 & 75.0 \\
    Nemotron-3-Embed-8B$^{\dagger}$ & 73.6 & 74.7 & 72.2 & 73.9 & 78.5 & 67.7 & 75.0 & 75.4 & 67.2 & 73.0 \\
    \texttt{pplx-embed-v1-4b}$^{\dagger}$ & 65.9 & 68.7 & 71.0 & 69.0 & 71.7 & 62.6 & 71.1 & 68.7 & 63.5 & 68.8 \\
    \bottomrule
  \end{tabular}}
\end{table}

\vtwosize{9b} is the strongest model on every query set and on every metric except evidence recall, where Nemotron-3-Embed-8B~\citep{nemotron3embed_2026} leads. We suspect that evidence chunks have high lexical overlap with the queries, which favors a non-contextual model such as Nemotron-3-Embed-8B. Across the four query sets, it outperforms \vone{} by 11 to 17 points and voyage-context-4 by 2 to 3 points at Answer@10, and it reaches full coverage with the fewest tokens. The 0.6B model is 0.8 points behind voyage-context-4 at Answer@10 on the base queries, ahead of it on evidence and tokens to coverage, and 8.2 points ahead of \vone{}. Context helps in both model families. voyage-context-4 outperforms voyage-4-large by 9.4 points, and \vone{} outperforms its non-contextual base by 6.8 points at Answer@10. Rewording the queries lowers Answer@10 for every model, least when paraphrased and most for indirect queries, which replace named entities with descriptions. \vtwosize{9b} achieves the best results in all variants of the query. For comparability with other benchmarks, we also report nDCG@10 on the base queries, which ranks the models in the same order. It is 62.8\% for \vtwosize{9b}, 59.8\% for voyage-context-4, 57.9\% for \vtwosize{0.6b}, 45.1\% for \vonesize{4b}, 43.9\% for voyage-4-large, 42.0\% for Nemotron-3-Embed-8B, 38.2\% for \vonesize{0.6b} and 35.7\% for \texttt{pplx-embed-v1-4b}. Document@10 is close to saturation, with every contextual model at or above its non-contextual counterpart. In the single-chunk setting, which embeds each page as one vector (truncated to 32,768 tokens), Document@10 on \qdchunk{} stays within one point of ranking each page by its best chunk for both models.

Our analysis shows that the benefit of context depends on the kind of evidence. Context helps most where the evidence and the answer do not repeat the query's specific terms, such as names, numbers or years. Models without context remain competitive on evidence that restates such terms, and on ranking evidence within a page they have already retrieved. The advantage of context therefore lies mainly in finding the right pages and in surfacing their evidence from the whole corpus.

Table~\ref{tab:q2d-chunk-domain-results} breaks Answer@10 down by the ten most frequent domains, which cover 89\% of the queries (Appendix~\ref{app:q2d-chunk-domains}). \vtwosize{9b} leads in every domain, and the 0.6B model stays within 3.0 points of voyage-context-4 in each of them, ahead of it in health, business and science.

 \paragraph{ConTEB.} Table~\ref{tab:conteb-results} reports the results on ConTEB, on which our distillation approach sets a new state of the art. \vtwosize{9b} reaches the highest average, 1.3 points above \vone{} and 3.0 points above voyage-context-4, and leads on five of the eight tasks. In contrast to \vone{}, which was trained on the ConTEB training set and leads on Insurance and NarrativeQA, \vtwo{} has not seen any ConTEB data. Nemotron-3-Embed-8B scores highest on COVID-QA. As \citet{conteb} point out, its query-chunk pairs are largely extractive and match on technical medical terms, so document context is less useful. On this task, both non-contextual models outperform all contextual ones. \vtwosize{0.6b} is 0.3 points above voyage-context-4 on average. On Insurance and Football, where chunks need their document context~\citep{conteb}, the non-contextual models trail the contextual ones by at least 65 and 24 points, respectively.

\begin{table}[t]
  \centering
  \small
  \caption{Task-level nDCG@10 on ConTEB in the query-to-chunk setting. The average is the unweighted mean across the eight tasks; best value per column in bold. Scores are in \%. $^{\dagger}$Non-contextual model, which encodes every chunk independently.}
  \label{tab:conteb-results}
  \resizebox{\textwidth}{!}{%
  \begin{tabular}{lccccccccc}
    \toprule
    Model & Avg. & Geography & ESG & Insurance & Football & SQuAD & NarrativeQA & MLDR & COVID-QA \\
    \midrule
    \vtwosize{9b} & \textbf{83.4} & \textbf{94.3} & \textbf{67.8} & 99.1 & \textbf{81.6} & \textbf{88.7} & 84.8 & \textbf{90.1} & 60.4 \\
    \vtwosize{0.6b} & 80.7 & 94.1 & 67.3 & 88.0 & 79.2 & 87.3 & 83.4 & 89.8 & 56.1 \\
    voyage-context-4 & 80.4 & 93.9 & 57.4 & 99.7 & 81.3 & 84.3 & 82.5 & 87.6 & 56.6 \\
    \vonesize{4b} & 82.1 & 93.5 & 63.4 & \textbf{100.0} & 78.1 & 83.7 & \textbf{86.9} & 89.4 & 62.0 \\
    voyage-4-large$^{\dagger}$ & 67.0 & 86.9 & 61.6 & 22.6 & 53.5 & 78.5 & 83.5 & 83.8 & 65.2 \\
    Nemotron-3-Embed-8B$^{\dagger}$ & 63.8 & 80.7 & 51.6 & 21.9 & 39.0 & 80.5 & 83.0 & 81.8 & \textbf{72.0} \\
    \bottomrule
  \end{tabular}}
\end{table}

\subsection{Document retrieval}
\label{sec:document-level-retrieval}

The previous subsection was concerned with the evaluation of chunk-level retrieval performance. At the document level, we measure the retrieval performance via the markdown version of the public split of ViDoRe-V3~\citep{vidorev3}. We evaluate our models with 128-token (multi) and whole-document (single) chunking. In the multi-chunk setting, we reuse the chunking strategy from Q2D-Chunk, with the chunk size set to 128 tokens, and rank documents according to their highest-scoring chunk. In the single-chunk setting, we embed the whole document by interpreting it as a single chunk. We evaluate the corresponding setups for our baselines and report all results in Tables~\ref{tab:vidore-markdown-single} and~\ref{tab:vidore-markdown-128} of Appendix~\ref{app:vidore-markdown}. The results are summarized in Table~\ref{tab:vidore-markdown}, where we report both settings for our models and the best-performing setting (according to average nDCG@10) for the baselines.
We find that our 9B contextual model with 128-token chunking is only outperformed by voyage-context-4. When evaluated with a single chunk, it is also outperformed by voyage-4-large and Nemotron-3-Embed-8B.
In the sub-billion size class, our 0.6B model is the strongest in both settings, outperforming voyage-4-nano~\citep{voyage4_2026}, EmbeddingGemma~2~\citep{embeddinggemma2_2026}, and Qwen3-Embedding-0.6B~\citep{qwen3-embedding}.
\begin{table}[t]
  \centering
  \fontsize{8.5}{11}\selectfont
  \renewcommand{\arraystretch}{1.0}
  \setlength{\tabcolsep}{1.5pt}
  \caption{Text-to-Markdown retrieval on ViDoRe-V3. Scores are nDCG@10 (\%), averaged equally over six languages per subtask; Avg.\ is the mean over eight subtasks.}
  \label{tab:vidore-markdown}
  \begin{tabular*}{\textwidth}{@{\extracolsep{\fill}}lcrrrrrrrrr@{}}
    \toprule
    \textbf{Model} & \textbf{Chunk} & \textbf{Avg.} & Fin(En) & Fin(Fr) & Indust. & Pharma & CS & HR & Energy & Physics \\
    \midrule
    \multicolumn{11}{@{}l}{\textbf{$\geq$1B parameters or undisclosed}} \\
    \addlinespace[2pt]
    \vtwosize{9b} & multi & \underline{61.18} & \textbf{64.19} & \textbf{51.06} & 49.33 & 65.42 & \underline{77.85} & \underline{62.75} & \underline{69.58} & 49.28 \\
    \vtwosize{9b} & single & 58.25 & 59.91 & 47.05 & 46.27 & 65.43 & 75.10 & 57.81 & 65.42 & 49.02 \\
    voyage-context-4 & multi & \textbf{62.76} & \underline{63.68} & \underline{50.24} & \textbf{52.64} & \textbf{68.74} & \textbf{80.65} & \textbf{65.83} & \textbf{70.60} & \underline{49.72} \\
    Nemotron-3-Embed-8B$^{\dagger}$ & single & 60.63 & 62.24 & 47.17 & \underline{50.88} & 67.19 & 77.08 & 62.21 & 67.18 & \textbf{51.11} \\
    voyage-4-large$^{\dagger}$ & single & 59.63 & 61.14 & 47.20 & 48.64 & \underline{67.67} & 76.16 & 61.29 & 66.39 & 48.52 \\
    \midrule
    \multicolumn{11}{@{}l}{\textbf{$<$1B parameters}} \\
    \addlinespace[2pt]
    \vtwosize{0.6b} & multi & \textbf{58.51} & \textbf{62.52} & \textbf{47.56} & \textbf{45.26} & \textbf{63.96} & \textbf{73.40} & \textbf{60.73} & \textbf{67.83} & \textbf{46.79} \\
    \vtwosize{0.6b} & single & \underline{55.81} & \underline{59.14} & \underline{43.51} & \underline{43.47} & \underline{63.37} & 70.69 & \underline{56.22} & \underline{63.62} & \underline{46.47} \\
    voyage-4-nano$^{\dagger}$ & single & 53.34 & 56.24 & 39.64 & 36.48 & 60.95 & \underline{72.06} & 55.02 & 61.43 & 44.86 \\
    EmbeddingGemma 2$^{\dagger}$ & single & 47.63 & 50.91 & 33.69 & 36.99 & 53.53 & 61.43 & 49.12 & 54.94 & 40.45 \\
    Qwen3-Embedding-0.6B$^{\dagger}$ & multi & 45.82 & 43.28 & 26.05 & 35.31 & 51.97 & 68.60 & 45.67 & 53.67 & 42.02 \\
    \bottomrule
  \end{tabular*}
  \par\vspace{4pt}
  \begin{minipage}{\textwidth}
    \fontsize{8}{10}\selectfont
    \textit{Chunk:} multi = 128-token chunks; single = one chunk per page. Baselines use their configuration with the highest Avg.; our models show both. Within each size group, \textbf{bold} and \underline{underlining} mark the best and second-best. $^{\dagger}$Non-contextual model.
  \end{minipage}
\end{table}

\subsection{Retrieval quality and storage costs}
\begin{figure}[t]
  \centering
  \includegraphics[width=\linewidth]{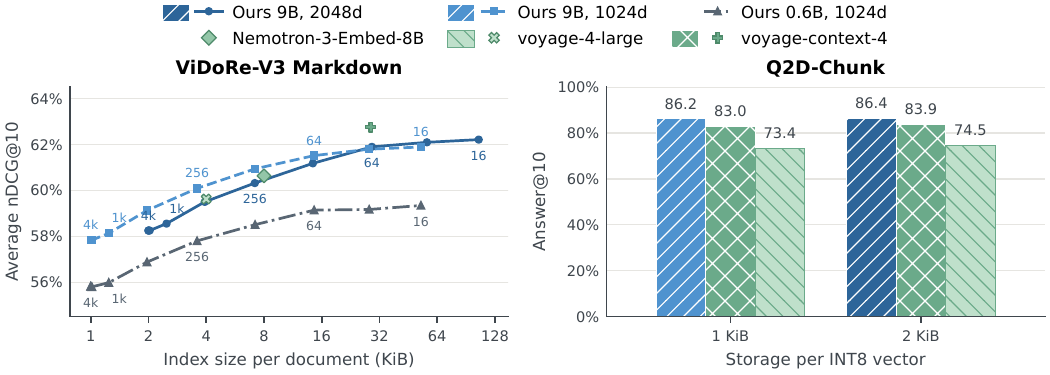}
  \caption{Retrieval quality versus index size. Left: mean nDCG@10 on ViDoRe-V3 Markdown across eight public tasks versus average index size per document. Our curves use one byte per embedding dimension; baseline points assume FP16 vectors. Nemotron-3-Embed-8B and voyage-4-large use one chunk per document at 4096 and 2048 dimensions, respectively; voyage-context-4 uses 128-token chunks at 2048 dimensions. Right: Q2D-Chunk performance for 1 and 2 KiB INT8 vectors for our 9B model and Voyage, assuming Voyage INT8 matches FP32.}
  \label{fig:vidore-markdown-chunk-sensitivity}
\end{figure}

A contextual embedding model produces one vector per chunk, so an index built with our models contains exactly as many vectors as a conventional chunk-level index built with a non-contextual model using the same chunking strategy. The storage footprint is governed by the size of embedding vectors. Our models produce efficient int8-quantized embeddings that may be truncated from 2048 to 1024 dimensions for the 9B model. We study how index size impacts retrieval quality in Figure~\ref{fig:vidore-markdown-chunk-sensitivity}. In the left subplot, we illustrate how document-level retrieval performance on ViDoRe-V3 markdown changes as the chunking granularity and embedding dimension are varied. We find that performance improves with smaller chunks and plateaus as the chunk size exceeds the typical length of documents within the corpus. Finer chunking in combination with truncation to 1024 dimensions usually leads to improved performance at the same index size.
In the right subplot, we investigate how chunk-level performance on Q2D-Chunk changes with truncation. 

\subsection{Visual document retrieval}
\label{sec:vdr-results}
\begin{figure}
    \centering
    \includegraphics[width=\linewidth]{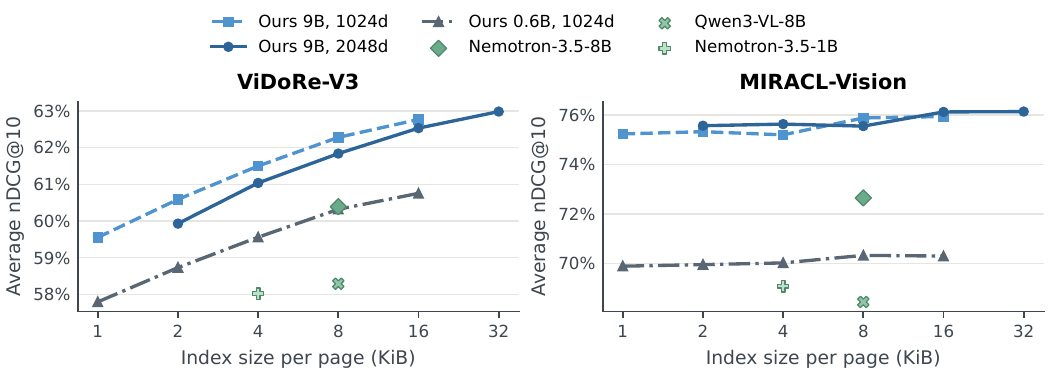}
    \caption{Average nDCG@10 versus index size per page on the eight public
    ViDoRe-V3 tasks (left) and MIRACL-Vision, averaged across 18 languages (right).
    Our curves use one byte per embedding dimension and 1, 2, 4, 8, or 16 chunks
    per page. Dense baselines assume one FP16 vector per page.}
    \label{fig:visual-retrieval}
\end{figure}

We observe that our contextual models exhibit strong VDR capabilities, despite the 9B model being trained only on text datasets. We recall from Subsection~\ref{sec:image-chunking-method} that we chunk images by dividing them into horizontal bands and inserting separation markers at the chunk interfaces before the forward pass through the text tower.
We evaluate our models on the public split of ViDoRe-V3~\citep{vidorev3} and on MIRACL-Vision~\citep{miracl-vision}. Since chunking allows page-level retrieval using standard ANN implementations for dense embedding models, we compare to leading dense baselines for VDR: Qwen3-VL-8B~\citep{qwen3-vl-embedding} and Nemotron-3.5-1B and Nemotron-3.5-8B~\citep{nemotron35_2026}. In Figure~\ref{fig:visual-retrieval}, we illustrate how performance behaves as the number of chunks is varied from 1 chunk to up to 16 chunks per image.
Generally, we observe that performance scales with index size on ViDoRe-V3 while chunking has little effect on MIRACL-Vision.
On ViDoRe-V3, our 9B model outperforms Qwen3-VL-8B and Nemotron-3.5-1B with a single chunk and substantially smaller index size. When using two chunks per image and 1024-dimensional embeddings, our 9B model also outperforms Nemotron-3.5-8B while using a $4\times$ smaller index.
Using two chunks per image, our 0.6B model beats Qwen3-VL-8B and Nemotron-3.5-1B while using a smaller index. To outperform Nemotron-3.5-8B, however, our 0.6B model needs double the index size. 
On MIRACL-Vision, performance varies little with chunking, indicating that chunking may be beneficial in some, but not all, settings.
Our 0.6B and 9B models outperform Qwen3-VL-8B and Nemotron-3.5-1B across all chunk settings while 9B also outperforms Nemotron-3.5-8B across all chunk settings.

These results demonstrate that our contextual models are effective visual document retrievers at the page level. We also investigate whether they are effective retrievers at the chunk level. To this end, we take ground-truth positive query-page pairs from ViDoRe-V3, embed 8 chunks per page, and check which chunk is ranked highest by our models. We present uncurated examples for the 9B model in Appendix~\ref{app:vdr-examples}. Qualitatively, we observe that the top-ranked chunks tend to contain the relevant information. We quantify this observation in Appendix~\ref{app:vdr-localization} by highlighting the top-ranked chunk on the page and letting a decision model~\citep{pplx-decider-v1.2} judge whether the highlighted band contains the relevant information. We compare this to a control in which the highlighted band is chosen randomly from chunk rankings on a held-out collection of query-document pairs. We find that our contextual models achieve substantially higher agreement (72.1\% for 9B, 69.8\% for 0.6B) with the decision model than the control (25.1\%).
Similar emergent localization behavior has previously been observed in ColPali~\citep{colpali}. Ablations of the image-chunking components are reported in Appendix~\ref{app:vdr-ablations}.

%% file: core/6_ablations.tex
\section{Ablations}
\label{sec:ablations}

Our objective combines a chunk-level distillation loss with a document-level contrastive loss (Section~\ref{sec:objective}). To isolate the contribution of each term, we train three 9B models that differ only in the objective: the contrastive loss alone, the distillation loss alone, and their combination. All three start from the same backbone and share the batch size and learning-rate schedule, and are compared at the same training step, 687 of 1,600 optimizer steps, within the stable phase of the schedule. Table~\ref{tab:ablation-loss} reports the performance on chunk-level (Q2D-Chunk and ConTEB) and document-level (ViDoRe-V3 Markdown with 128-token chunks) retrieval benchmarks.

The distillation loss is necessary for chunk retrieval: without it, Q2D-Chunk drops by 15.2 points on average and ConTEB drops by 14.6 points. Distillation alone preserves most of the document-level quality, due to the document-level term contained in its batch-wide KL (Eq.~\eqref{eq:kl-decomposition}). The contrastive term still improves all three benchmarks.

\begin{table}[H]
  \centering
  \small
  \caption{Training-objective ablation for the 9B model. Scores in \%; best per benchmark in bold.}
  \label{tab:ablation-loss}
  \setlength{\tabcolsep}{5pt}
  \begin{tabular*}{\textwidth}{@{\extracolsep{\fill}}llccc@{}}
    \toprule
    Benchmark & Metric & Contrastive only & Distillation only & Both (ours) \\
    \midrule
    Q2D-Chunk & Answer@10 & 71.1 & 85.8 & \textbf{86.3} \\
    ConTEB & nDCG@10 & 68.1 & 82.4 & \textbf{82.7} \\
    ViDoRe-V3 Markdown & nDCG@10 & 58.7 & 60.3 & \textbf{60.9} \\
    \bottomrule
  \end{tabular*}
\end{table}

%% file: core/7_conclusion.tex
\section{Conclusion}
\label{sec:conclusion}

We presented \vtwo{}, a family of contextual embedding models trained by distilling graded chunk-level relevance from a query-aware context compression model instead of gold-chunk labels. To address the lack of benchmarks for queries whose answers span multiple chunks or require supporting evidence, we introduced \contextbench{} and \qdchunk{}, which label both answer and evidence chunks. \vtwosize{9b} sets a new state of the art on both benchmarks and on ConTEB. Both models are also state-of-the-art dense visual document retrievers on ViDoRe-V3, with further gains when each page is embedded as several chunks. Evidence retrieval remains challenging on both benchmarks. We publish \qdchunk{} as a public leaderboard.

\clearpage
\section*{Ethics and Privacy Statement} 
\qdchunk{} is derived from production search traffic of a commercial conversational search assistant, and its corpus consists of publicly accessible web pages. Use of this traffic is consistent with the applicable user agreements, privacy policy, and data protection law. We exclude searches from users who have opted out of the use of their data for model improvement or from analytics, as well as logged-out, deleted, likely fraudulent and sensitive-domain searches. Users do not author the benchmark queries: each query starts from a search query that the assistant's agent generated from the user's request, which an LLM then rewrites into a natural-language question; the paraphrased, indirect, and scenario variants are further LLM rewrites. We apply PII detection to every query and exclude any query flagged as containing PII. In the retrieved pages, detected personal information is replaced with typed placeholders, and pages whose answer spans contain a placeholder are excluded. Relevance and evidence labels are produced by LLM labelers and an LLM adjudicator, not by human annotators, and should be read as model-assigned judgments. \qdchunk{} remains a private benchmark evaluated through a hosted leaderboard; only aggregate results are shared externally.

%% file: core/8_appendix.tex
\appendix

\section{Training hyperparameters}
\label{app:hyperparams}

\begin{table}[H]
  \centering
  \small
  \caption{Training hyperparameters of the 9B model, trained in a single stage with Matryoshka and int8 quantization-aware training from the first step.}
  \label{tab:hyperparams}
  \begin{tabular}{ll}
    \toprule
    Steps & 1600 \\
    Warmup / stable / decay steps & 300 / 700 / 600 \\
    Learning-rate schedule & warmup--stable--decay \\
    Peak learning rate & $1.25 \times 10^{-4}$ \\
    Optimizer & AdamW, weight decay 0.01 \\
    Optimizer batch (pairs) & 65{,}536 \\
    In-batch negatives per query & 16{,}384 \\
    LoRA rank / $\alpha$ & 128 / 256 \\
    Temperatures $\tau = \tau^{\mathrm{S}}$ / $\tau^{\mathrm{T}}$ & 0.02 / 0.075 \\
    Teacher top-$k$ tokens & 16 \\
    Max.\ training document length & 1000 tokens \\
    Precision / parallelism & bf16 AMP, hybrid-sharded FSDP, gradient caching \\
    \bottomrule
  \end{tabular}
\end{table}

\section{Visual document chunking}
\label{app:visual-chunking}
\begin{figure}[htbp]
  \centering
  \resizebox{\linewidth}{!}{\input{figures/visual_chunking}}
  \caption{Visual document chunking, illustrated with four horizontal bands. The soft visual token sequence is passed through the text tower.}
  \label{fig:visual-chunking}
\end{figure}
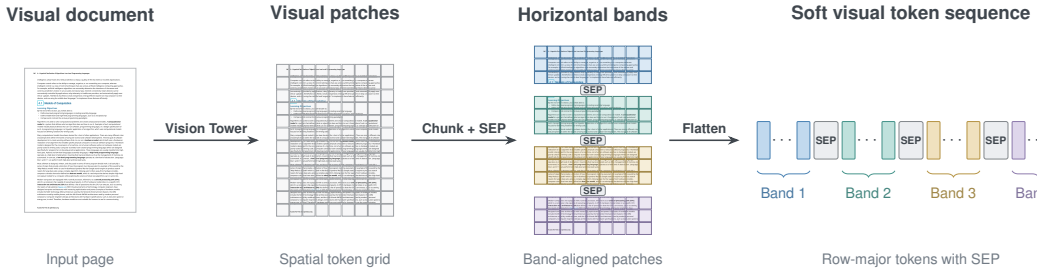

In Figure~\ref{fig:visual-chunking}, we illustrate how we chunk pages of visual documents. The vision tower encodes the page into soft visual tokens, first shown as an ungrouped spatial patch grid, then partitioned into colored bands. Flattening preserves row-major order, with a learned \texttt{<|chunk\_sep|>} token (SEP) inserted between adjacent band spans. We choose MRoPE coordinates for the separator tokens to place them at the band interfaces: Let $b$ be the index of the first image token, and $W$ be the width of the spatial visual token grid. For the separation token placed before the row $s$, we set its MRoPE coordinates to $\left(b, b + s - 0.5, b + \frac{W -1}{2}\right)$.

\section{Context dependency types in \contextbench{}}
\label{app:context-bench-capabilities}

Table~\ref{tab:context-bench-capabilities} lists the twelve types of context dependency tested by \contextbench{}, each with an illustrative query, its answer chunk and the document context that the answer requires.

\begin{table}[p]
  \centering
  \footnotesize
  \renewcommand{\arraystretch}{1.25}
  \begin{tabular}{p{0.15\linewidth}p{0.22\linewidth}p{0.2\linewidth}p{0.31\linewidth}}
    \toprule
    Type & Example query & Answer chunk & Required document context \\
    \midrule
    1. Entity identity & How much did Northlake spend on share repurchases? & The company repurchased \$12.8 billion of shares. & The introduction identifies Northlake as ``the company.'' A similar Southlake report is not relevant. \\
    2. Period or version & Which cartridge fits the M40 revision B? & Install the KC-42 cartridge. & The applicability section says these instructions cover revision B, not revision C. \\
    3. Definitions and aliases & What withdrawal rate did the 15 mg group have? & Zeta had a withdrawal rate of 12.7\%. & The report defines Zeta as the 15 mg group. Without that definition, the chunk does not identify the requested group. \\
    4. Table structure & What is the Max model's battery capacity? & Typical capacity: 4,250 \textbar{} 4,700 \textbar{} 5,150. & The table header establishes Mini \textbar{} Base \textbar{} Max, measured in mAh. The answer is therefore 5,150 mAh. \\
    5. Conditions and exceptions & How long is battery coverage for commercial use? & Battery coverage lasts 18 months. & A section introduction restricts this coverage to commercial installations; household installations have a different warranty. \\
    6. References and pronouns & When did Mara become deputy mayor? & She took office in 2021. & The preceding narrative establishes that ``she'' refers to Mara, not Elin, who is also discussed. \\
    7. List order and position & Which song opened the encore? & 14. Lantern Road. & A note elsewhere says songs 14--16 constituted the encore. Track 14 is therefore its opening song. \\
    8. Section scope & Which lens was used for the film's cinematography? & A 35 mm prime was used throughout. & This passage belongs to ``Principal Photography,'' not the separate section about publicity photographs. \\
    9. Speaker attribution & What reopening date did the mayor promise? & We will reopen on June 12. & The transcript identifies the mayor as the speaker. The same words from a contractor would not establish the mayor's promise. \\
    10. Changes over time & What notice period applied after the May amendment? & The notice period is now 45 days. & The amendment history establishes that this change took effect in May, replacing the earlier 30-day rule. \\
    11. Causal relationships & Why did the west pump stop? & That fault caused the shutdown. & The investigation identifies ``that fault'' as a seized bearing in the west pump, and distinguishes it from an unrelated inlet obstruction. \\
    12. Figurative or cross-language meaning & \emph{Figurative.} What does ``the winter room'' symbolize in the poem? \newline \emph{Cross-language.} Which component does the French instruction tell the technician to remove? & \emph{Figurative.} It represents exile, rather than a physical room. \newline \emph{Cross-language.} Retirez le capot. (Remove the cover.) & \emph{Figurative.} The commentary identifies ``it'' as the poem's recurring ``winter room'' image. \newline \emph{Cross-language.} The guide's glossary specifies that ``capot'' means the protective motor hood, not the shipping cover. \\
    \bottomrule
  \end{tabular}
  \caption{Twelve types of context dependency tested by \contextbench{}, with illustrative examples.}
  \label{tab:context-bench-capabilities}
\end{table}

\clearpage
\section{\qdchunk{} query variants}
\label{app:q2d-chunk-variants}

Figure~\ref{fig:q2d-chunk-variants} shows two \qdchunk{} queries as base queries and in the three reworded variants. The paraphrased variant shares fewer keywords with the answer chunks, the indirect variant replaces named entities with descriptions that identify them uniquely, and the scenario variant is a first-person forum post, with a title and a body, expressing the same need.

\begin{figure}[H]
  \centering
  \begin{tcolorbox}[colback=white, colframe=black!20, boxrule=0.5pt, arc=2pt, left=6pt, right=6pt, top=4pt, bottom=4pt]
    \small
    \textbf{Finance.}\\[2pt]
    \textbf{Base.} What are the official 2026 IRS standard deduction for single filers, 403(b) contribution limit, and student loan interest deduction?\\[1pt]
    \textbf{Paraphrased.} According to the IRS, what are the 2026 amounts for the standard deduction for an individual filing alone, the 403(b) contribution limit, and the student loan interest deduction?\\[1pt]
    \textbf{Indirect.} What are the official 2026 standard deduction for single filers set by the federal tax-collecting agency, the contribution limit for the tax-advantaged employer-sponsored retirement savings arrangement for employees of certain tax-exempt organizations and public schools, and the student loan interest deduction?\\[1pt]
    \textbf{Scenario.} \emph{What are the official 2026 IRS amounts for these tax items?} I'm preparing my tax and retirement planning and need to know the 2026 IRS standard deduction for single filers, the 403(b) contribution limit, and the student loan interest deduction.
  \end{tcolorbox}
  \vspace{2pt}
  \begin{tcolorbox}[colback=white, colframe=black!20, boxrule=0.5pt, arc=2pt, left=6pt, right=6pt, top=4pt, bottom=4pt]
    \small
    \textbf{Health.}\\[2pt]
    \textbf{Base.} What are the co-occurring traits and diagnostic considerations for ADHD and autism spectrum disorder in children?\\[1pt]
    \textbf{Paraphrased.} In children, what features may occur together with ADHD and autism spectrum disorder, and what factors should clinicians consider when evaluating these conditions?\\[1pt]
    \textbf{Indirect.} What traits commonly occur together, and what diagnostic considerations apply, when evaluating children for the neurodevelopmental disorder involving persistent inattention and/or hyperactivity-impulsivity and the neurodevelopmental condition characterized by social-communication differences and restricted or repetitive behaviors?\\[1pt]
    \textbf{Scenario.} \emph{What traits and diagnostic considerations overlap between ADHD and autism spectrum disorder?} I'm trying to understand how ADHD and autism spectrum disorder may appear together in a young person. What co-occurring traits should I look for, and what diagnostic considerations are important when evaluating both conditions?
  \end{tcolorbox}
  \caption{\qdchunk{} queries in their four wordings. The scenario variant is shown as its post title (italic) followed by its body.}
  \label{fig:q2d-chunk-variants}
\end{figure}

\section{A \qdchunk{} example}
\label{app:pplx-conteb-example}

Figure~\ref{fig:pplx-conteb-example} shows one \qdchunk{} query with its answer and evidence groups and an excerpt of one of its pages, labeled as 128-token chunks.

\begin{figure}[H]
  \centering
  \begin{tcolorbox}[colback=white, colframe=black!20, boxrule=0.5pt, arc=2pt, left=6pt, right=6pt, top=5pt, bottom=5pt]
    \small
    \textbf{Query.} What are the Asian Development Bank's GDP growth forecasts for Kazakhstan, China, and the Caucasus and
    Central and West Asia in 2026 and 2027?\\[1pt]
    \textbf{Groups.} \textbf{A1} Kazakhstan forecasts \quad \textbf{A2} China forecasts \quad \textbf{A3} Caucasus and
    Central and West Asia forecasts \quad \textbf{E1} the ADB \emph{Asian Development Outlook}, April 2026\par
    \vspace{4pt}
    \begin{pcchunk}{pcevidence}{c0 \textbullet{} evidence E1}
      Economic Forecasts: Asian Development Outlook April 2026 \textbar{} Asian Development Bank \ldots{}
      \textbf{Economic forecasts have since been updated} \ldots{} Economic Forecasts for Asia and the Pacific: April
      2026 \ldots
    \end{pcchunk}
    \begin{pcchunk}{pcother}{c1}
      This brief sets out updated estimates of the potential impact of the 2026 Middle East conflict on developing Asia
      and the Pacific \ldots
    \end{pcchunk}
    \pcgap
    \begin{pcchunk}{pcanswer}{c22 \textbullet{} answer A3}
      Several factors are projected to moderate growth in the Caucasus and Central and West Asia. Expansion is expected
      to decline from 4.6\% in 2025 to 4.2\% in 2026 and recover slightly to 4.4\% in 2027. This reflects mainly a
      slowdown in Kazakhstan as oil output plateaus near capacity \ldots
    \end{pcchunk}
    \begin{pcchunk}{pcother}{c23}
      Growth in Georgia and Tajikistan is also expected to slow as weaker global demand weighs on exports \ldots
    \end{pcchunk}
    \pcgap
    \begin{pcchunk}{pcanswer}{c26 \textbullet{} answer A2}
      \ldots{} Developing East Asia. Growth in the PRC is projected to slow, reflecting subdued private consumption and
      weaker external demand. GDP growth is expected to decline from 5.0\% in 2025 to 4.6\% in 2026 and 4.5\% in 2027.
    \end{pcchunk}
  \end{tcolorbox}
  \caption{An example from \qdchunk{}: a query, its answer and evidence groups, and an excerpt
  of one of its pages as 128-token chunks (\textcolor{pcanswer}{\rule{6pt}{6pt}}~answer,
  \textcolor{pcevidence}{\rule{6pt}{6pt}}~evidence, \textcolor{pcother}{\rule{6pt}{6pt}}~other; text shortened). The
  answer chunks state the forecasts but not whose forecasts they are or from which edition; only the evidence chunk,
  the page title, ties them to the ADB's April 2026 outlook, which matters because the page notes that the forecasts
  have since been updated. The page gives no figure for Kazakhstan alone, so A1 has no chunk here.}
  \label{fig:pplx-conteb-example}
\end{figure}

\section{\qdchunk{} domains and languages}
\label{app:q2d-chunk-domains}

\begin{figure}[p]
  \centering
  \includegraphics[width=\textwidth]{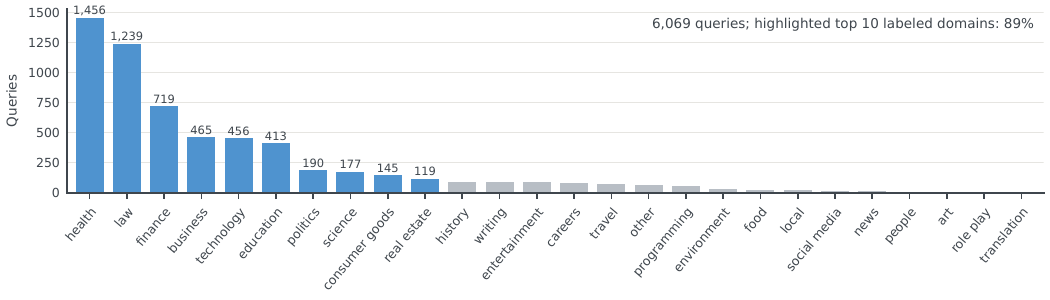}
  \caption{Queries per top-level domain in \qdchunk{} (base queries, 6,069 queries). Highlighted: the ten most frequent domains, reported in Table~\ref{tab:q2d-chunk-domain-results}.}
  \label{fig:q2d-chunk-domains}
\end{figure}

Figure~\ref{fig:q2d-chunk-domains} shows the number of \qdchunk{} queries per top-level domain on the base queries. Health, law and finance account for 56\% of the queries. The ten most frequent domains, highlighted, cover 89\% of the queries, and Table~\ref{tab:q2d-chunk-domain-results} reports Answer@10 on each of them. The reworded query sets follow a similar distribution.

\begin{figure}[p]
  \centering
  \includegraphics[width=\textwidth]{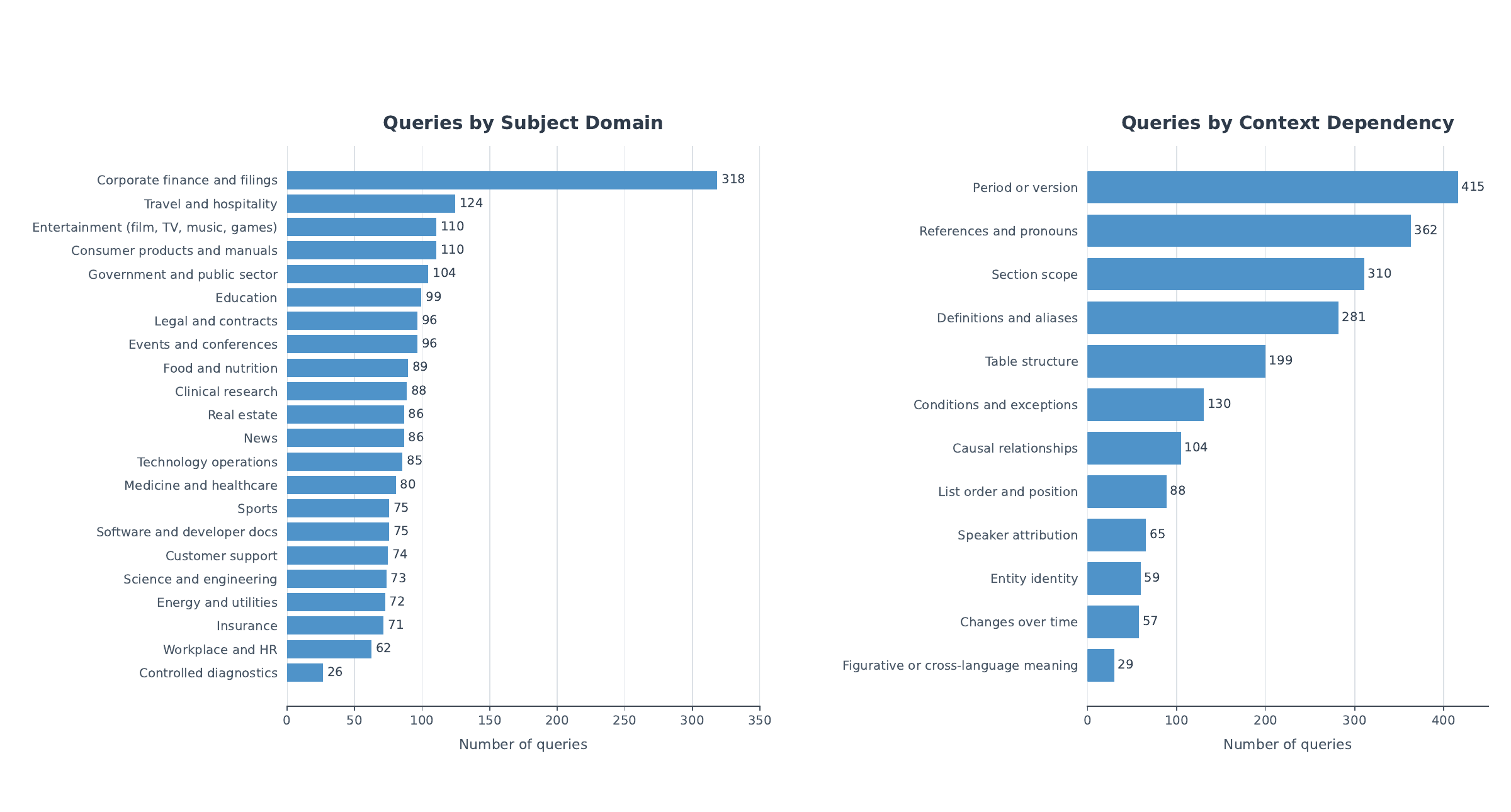}
  \caption{Context-bench query coverage by subject domain and type of context dependency.}
  \label{fig:context-bench-query-coverage}
\end{figure}

\begin{figure}[p]
  \centering
  \includegraphics[width=\textwidth]{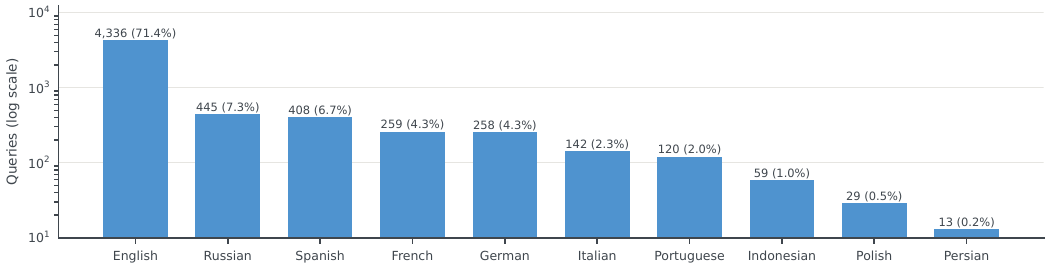}
  \caption{Queries per query language in \qdchunk{} (base queries, 6,069 queries; log scale).}
  \label{fig:q2d-chunk-languages}
\end{figure}

\section{Retrieval quality across domains and chunk-size sensitivity}
\label{app:q2d}

Beyond benchmarks designed for contextual embeddings, we assess the domain-specific performance of our model on MTEB-based query-to-chunk (Table~\ref{tab:domain-results}) and query-to-document (Table~\ref{tab:q2d-results}) retrieval. Our MTEB-based query-to-document suite (MTEB-Q2D) combines MTEB tasks such as LegalBench, FinanceBench, and HumanEval with additional retrieval tasks we curated from public datasets and scores whether the relevant document is retrieved. The query-to-chunk suite (MTEB-Q2C) is built from the same tasks by labeling, within each relevant document, which chunks are relevant to the query: a chunk is labeled relevant if it contains the exact answer span when the task provides one, and otherwise if a large language model judges it relevant to the query.

As can be seen, our model achieves the highest average score on chunk retrieval, with the clearest gains in the legal, technical and health domains. On document retrieval (Table~\ref{tab:q2d-results}), its performance is comparable to that of the strongest dense and contextual baselines. \vone{} is markedly weaker on both suites, which is consistent with it having been trained on the ConTEB training set: supervision tied to one benchmark's gold chunks transfers poorly to other domains, whereas our model draws its within-document supervision from the compression model over a broad training mixture.

\begin{figure}[H]
  \centering
  \begin{subfigure}[b]{0.48\textwidth}
    \centering
    \includegraphics[width=\linewidth]{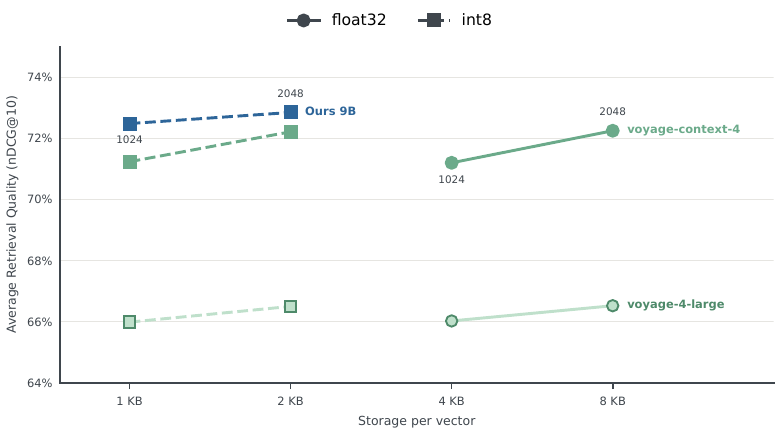}
    \caption{Retrieval quality vs.\ storage per vector (MTEB-Q2C).}
    \label{fig:storage-q2c}
  \end{subfigure}\hfill
  \begin{subfigure}[b]{0.34\textwidth}
    \centering
    \includegraphics[width=\linewidth]{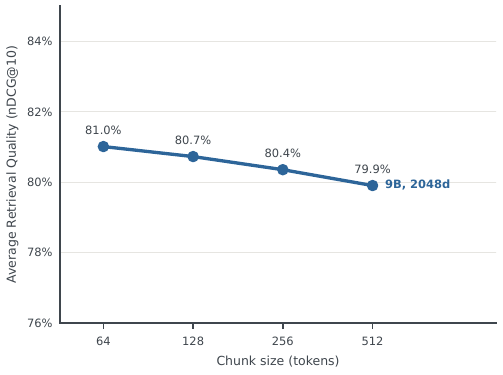}
    \caption{Chunk-size sensitivity (MTEB-Q2D).}
    \label{fig:chunk-sensitivity}
  \end{subfigure}
  \caption{(a) Mean nDCG@10 on the MTEB-Q2C tasks of Table~\ref{tab:domain-results} against storage per vector, in float32 (solid) and int8 (dashed), with labels giving the dimension. (b) Mean nDCG@10 on the MTEB-Q2D tasks of Table~\ref{tab:q2d-results}, with documents indexed at the given chunk size.}
  \label{fig:storage-chunk-appendix}
\end{figure}

Figure~\ref{fig:chunk-sensitivity} shows how retrieval quality on MTEB-Q2D changes with the chunk size used at indexing time for our 9B model.

\begin{table}[H]
  \centering
  \small
  \caption{Retrieval quality across domains on the MTEB-Q2C tasks. Mean nDCG@10 per domain, as unweighted means of the task-level scores in that domain; the average is the unweighted mean across all tasks. Best value per column in bold. Scores are in \%. $^{\dagger}$Non-contextual model, which encodes every chunk independently.}
  \label{tab:domain-results}
  \resizebox{\textwidth}{!}{%
  \begin{tabular}{lccccccc}
    \toprule
    Model & Avg. & Law & Tech & Finance & Health & Multilingual & Conversation \\
    \midrule
    \vtwosize{9b} & \textbf{72.9} & \textbf{68.2} & \textbf{48.0} & 81.6 & \textbf{83.2} & 90.4 & 72.4 \\
    voyage-context-4 & 72.3 & 65.2 & 46.5 & \textbf{82.9} & 82.2 & \textbf{90.9} & 73.2 \\
    \vonesize{4b} & 64.4 & 53.9 & 38.4 & 74.2 & 74.3 & 89.2 & 70.7 \\
    voyage-4-large$^{\dagger}$ & 66.5 & 56.0 & 37.0 & 79.1 & 77.6 & 88.2 & 73.1 \\
    Nemotron-3-Embed-8B$^{\dagger}$ & 66.6 & 55.2 & 37.5 & 77.1 & 81.8 & 89.3 & \textbf{74.9} \\
    \bottomrule
  \end{tabular}}
\end{table}

\begin{table}[H]
  \centering
  \small
  \caption{Retrieval quality across domains on the MTEB-Q2D tasks. Mean task score per domain, as unweighted means of the task-level main scores in that domain; the multilingual group also contains the remaining tasks, and the average is the unweighted mean across all tasks. Best value per column in bold. Scores are in \%. $^{\dagger}$Non-contextual model, which encodes every chunk independently.}
  \label{tab:q2d-results}
  \resizebox{\textwidth}{!}{%
  \begin{tabular}{lccccccc}
    \toprule
    Model & Avg. & Law & Tech & Finance & Health & Multilingual \& other & Conversation \\
    \midrule
    \vtwosize{9b} & 80.7 & 80.6 & \textbf{67.7} & 80.6 & 83.0 & 86.8 & 71.4 \\
    voyage-context-4 & \textbf{82.2} & \textbf{81.8} & 66.7 & \textbf{82.9} & \textbf{83.7} & \textbf{88.8} & 72.3 \\
    \vonesize{4b} & 77.0 & 76.6 & 61.9 & 77.7 & 78.0 & 82.4 & 70.5 \\
    voyage-4-large$^{\dagger}$ & 79.9 & 79.2 & 62.8 & 79.8 & 81.8 & 87.3 & 71.8 \\
    Nemotron-3-Embed-8B$^{\dagger}$ & 80.0 & 79.1 & 64.7 & 79.7 & 82.7 & 86.8 & \textbf{72.7} \\
    \bottomrule
  \end{tabular}}
\end{table}

\section{Document-level retrieval on ViDoRe-V3}
\label{app:vidore-markdown}

Tables~\ref{tab:vidore-markdown-single} and~\ref{tab:vidore-markdown-128} report the single-chunk and 128-token-chunk configurations, respectively, for all models in Table~\ref{tab:vidore-markdown}. Scores are nDCG@10 (\%), averaged equally over six query languages per subtask; Avg.\ is the unweighted mean over the eight subtasks. Within each size group, \textbf{bold} and \underline{underlining} mark the best and second-best scores in each column.

\begin{table}[H]
  \centering
  \fontsize{8.5}{11}\selectfont
  \renewcommand{\arraystretch}{1.0}
  \setlength{\tabcolsep}{1.5pt}
  \caption{Single-chunk text-to-Markdown retrieval on ViDoRe-V3, with one chunk per page. Scores are nDCG@10 in \%; -- denotes an unavailable result. $^{\dagger}$Non-contextual model, which encodes every chunk independently.}
  \label{tab:vidore-markdown-single}
  \begin{tabular*}{\textwidth}{@{\extracolsep{\fill}}lrrrrrrrrr@{}}
    \toprule
    \textbf{Model} & \textbf{Avg.} & Fin(En) & Fin(Fr) & Indust. & Pharma & CS & HR & Energy & Physics \\
    \midrule
    \multicolumn{10}{@{}l}{\textbf{$\geq$1B parameters or undisclosed}} \\
    \addlinespace[2pt]
    \vtwosize{9b} & 58.25 & 59.91 & 47.05 & 46.27 & 65.43 & 75.10 & 57.81 & 65.42 & \underline{49.02} \\
    Nemotron-3-Embed-8B$^{\dagger}$ & \textbf{60.63} & \textbf{62.24} & \underline{47.17} & \textbf{50.88} & \underline{67.19} & \textbf{77.08} & \textbf{62.21} & \textbf{67.18} & \textbf{51.11} \\
    voyage-4-large$^{\dagger}$ & \underline{59.63} & \underline{61.14} & \textbf{47.20} & \underline{48.64} & \textbf{67.67} & \underline{76.16} & \underline{61.29} & \underline{66.39} & 48.52 \\
    \midrule
    \multicolumn{10}{@{}l}{\textbf{$<$1B parameters}} \\
    \addlinespace[2pt]
    \vtwosize{0.6b} & \textbf{55.81} & \textbf{59.14} & \textbf{43.51} & \textbf{43.47} & \textbf{63.37} & \underline{70.69} & \textbf{56.22} & \textbf{63.62} & \textbf{46.47} \\
    voyage-4-nano$^{\dagger}$ & \underline{53.34} & \underline{56.24} & \underline{39.64} & 36.48 & \underline{60.95} & \textbf{72.06} & \underline{55.02} & \underline{61.43} & \underline{44.86} \\
    EmbeddingGemma 2$^{\dagger}$ & 47.63 & 50.91 & 33.69 & \underline{36.99} & 53.53 & 61.43 & 49.12 & 54.94 & 40.45 \\
    Qwen3-Embedding-0.6B$^{\dagger}$ & 44.43 & 44.44 & 25.27 & 32.39 & 51.92 & 65.62 & 40.21 & 51.68 & 43.89 \\
    \bottomrule
  \end{tabular*}
\end{table}

\begin{table}[H]
  \centering
  \fontsize{8.5}{11}\selectfont
  \renewcommand{\arraystretch}{1.0}
  \setlength{\tabcolsep}{1.5pt}
  \caption{Text-to-Markdown retrieval on ViDoRe-V3 with 128-token chunks. Scores are nDCG@10 in \%. $^{\dagger}$Non-contextual model, which encodes every chunk independently.}
  \label{tab:vidore-markdown-128}
  \begin{tabular*}{\textwidth}{@{\extracolsep{\fill}}lrrrrrrrrr@{}}
    \toprule
    \textbf{Model} & \textbf{Avg.} & Fin(En) & Fin(Fr) & Indust. & Pharma & CS & HR & Energy & Physics \\
    \midrule
    \multicolumn{10}{@{}l}{\textbf{$\geq$1B parameters or undisclosed}} \\
    \addlinespace[2pt]
    \vtwosize{9b} & \underline{61.18} & \textbf{64.19} & \textbf{51.06} & 49.33 & 65.42 & 77.85 & \underline{62.75} & \underline{69.58} & 49.28 \\
    voyage-context-4 & \textbf{62.76} & \underline{63.68} & \underline{50.24} & \textbf{52.64} & \textbf{68.74} & \textbf{80.65} & \textbf{65.83} & \textbf{70.60} & \textbf{49.72} \\
    Nemotron-3-Embed-8B$^{\dagger}$ & 58.23 & 56.25 & 40.45 & \underline{49.90} & 66.74 & \underline{78.18} & 59.52 & 65.29 & \underline{49.52} \\
    voyage-4-large$^{\dagger}$ & 57.75 & 54.40 & 39.86 & 48.41 & \underline{67.06} & 77.79 & 59.56 & 67.04 & 47.85 \\
    \midrule
    \multicolumn{10}{@{}l}{\textbf{$<$1B parameters}} \\
    \addlinespace[2pt]
    \vtwosize{0.6b} & \textbf{58.51} & \textbf{62.52} & \textbf{47.56} & \textbf{45.26} & \textbf{63.96} & \underline{73.40} & \textbf{60.73} & \textbf{67.83} & \textbf{46.79} \\
    voyage-4-nano$^{\dagger}$ & \underline{51.20} & \underline{49.40} & \underline{34.47} & 35.61 & \underline{59.13} & \textbf{73.44} & \underline{53.28} & \underline{61.13} & \underline{43.19} \\
    EmbeddingGemma 2$^{\dagger}$ & 46.11 & 44.93 & 29.36 & \underline{38.73} & 50.42 & 64.41 & 46.45 & 55.35 & 39.22 \\
    Qwen3-Embedding-0.6B$^{\dagger}$ & 45.82 & 43.28 & 26.05 & 35.31 & 51.97 & 68.60 & 45.67 & 53.67 & 42.02 \\
    \bottomrule
  \end{tabular*}
\end{table}

\section{Quantitative analysis of top chunk localization in ViDoRe-V3}
\label{app:vdr-localization}

\begin{figure}[H]
  \centering
  \includegraphics[width=\linewidth]{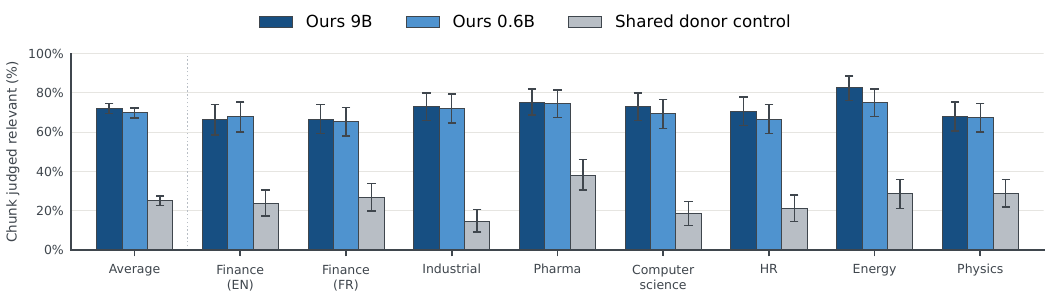}
  \caption{Chunk localization on ViDoRe-V3 for English queries using the 0.6B
  and 9B models and eight chunks per page. Bars show the proportion of highlighted
  regions judged relevant; Average is the unweighted mean across datasets.
  Error bars show pointwise 95\% percentile bootstrap intervals from matched
  query--page pairs resampled within each dataset.
  Interval overlap is not a paired significance test.}
  \label{fig:vdr-localization}
\end{figure}

We evaluate whether the highest-scoring chunk localizes query-relevant information within a ground-truth positive page.
We evaluate both the 0.6B and 9B models with eight chunks per page on the same 150 English query--page pairs from each of the eight public ViDoRe-V3 datasets (1,200 pairs in total).
Given the embedding model, we compute chunk-level embeddings for the given visual document and score the chunks against the query. We then select the top-ranking chunk, visually highlight it as in Appendix~\ref{app:vdr-examples} and present the highlighted query--document pair to a judge: \texttt{pplx-decider-v1.2}~\citep{pplx-decider-v1.2} assesses whether the highlighted region contains information that directly helps answer the query.
Both models share a control that highlights the band selected by a disjoint donor
query--page pair from the same dataset and language, using the 0.6B model's donor
bands. Figure~\ref{fig:vdr-localization} shows the rates at which the decider model judges the top chunk to be directly relevant:
the model-selected chunks average 69.83\% for 0.6B and 72.08\% for 9B, compared
with 25.08\% for the shared donor control. These correspond to improvements of
44.75 and 47.00 percentage points over the control, respectively.

\section{Ablations on visual document retrieval}
\label{app:vdr-ablations}
\begin{figure}[htbp]
    \centering
    \includegraphics[width=\linewidth]{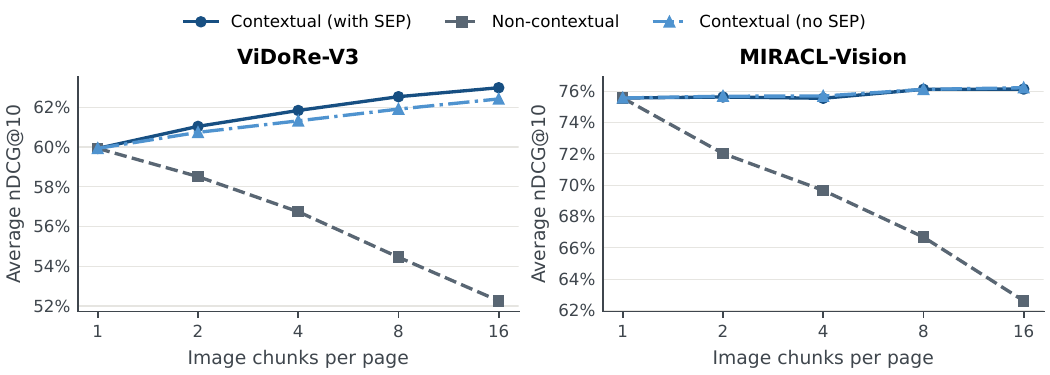}
    \caption{Ablation of contextual image chunking. We find that the joint (contextual) forward passes over chunks are important. Adding the separation token between chunks improves performance on ViDoRe-V3.}
    \label{fig:vdr_ablations}
\end{figure}

In Figure~\ref{fig:vdr_ablations} we show how different components of our image chunking method impact performance. 
We study a non-contextual baseline in which we chunk input images but perform independent forward passes for each chunk. In this baseline, performance degrades as the number of chunks increases. We conclude that contextual information is important and naive chunking alone does not improve VDR performance.
We also present an ablation variant in which we do not separate chunks by the learned separation token. We find that performance degrades on ViDoRe-V3 but barely changes on MIRACL-Vision. Hence, the separation token may be beneficial in some, but not all, settings.

\clearpage
\section{Examples of top chunks in ViDoRe-V3}
\label{app:vdr-examples}

{\fontsize{8}{9}\selectfont English-query examples grouped by ViDoRe-V3 dataset.
Yellow bands and red brackets mark the model-selected chunk on the ground-truth positive document. We show the first query-document pair for each dataset without manual selection.\par}

\begingroup
\newlength{\vdrImageHeight}
\setlength{\vdrImageHeight}{215pt}
\newcommand{\vdrExample}[3]{%
  \begin{minipage}[t]{\linewidth}
    \raggedright
    \begin{minipage}[t][44pt][t]{\linewidth}
      \fontsize{8.5}{10}\selectfont\textbf{Query.} #1
    \end{minipage}\par
    \parbox[c][\vdrImageHeight][c]{\linewidth}{%
      \centering\includegraphics[width=\linewidth,height=\vdrImageHeight,
        trim=#3,clip,keepaspectratio]{#2}%
    }\par
  \end{minipage}%
}

\begin{figure}[H]
  \centering
  \begin{minipage}[t]{0.485\linewidth}
    \raggedright
    {\fontsize{9.5}{11}\selectfont\bfseries\texttt{Vidore3ComputerScienceRetrieval}\par}
    \vspace{3pt}
    \vdrExample
      {What are the key differences between low-level and high-level programming languages regarding hardware interaction, execution speed, and developer productivity in computational systems?}
      {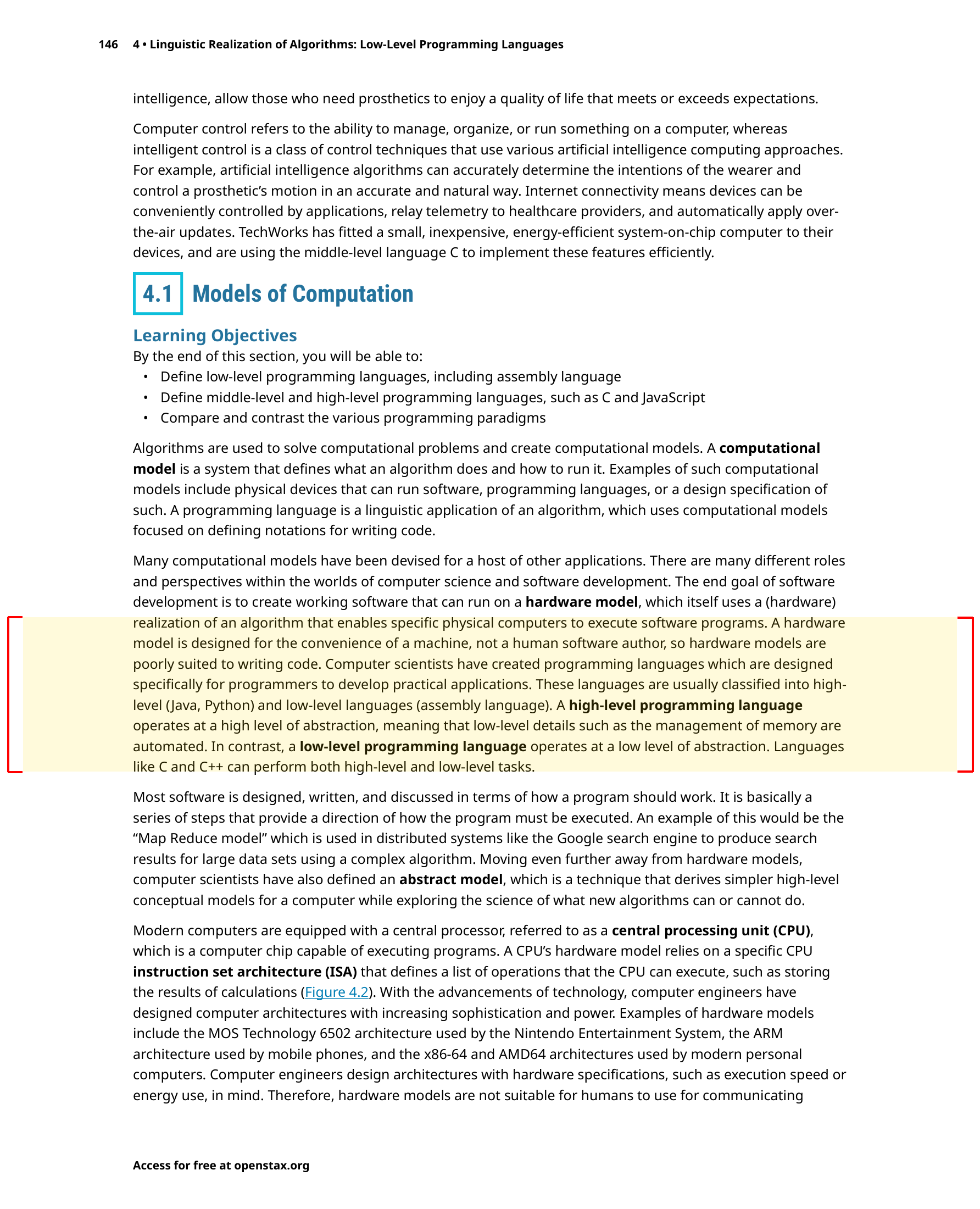}{0bp 51bp 0bp 56bp}
  \end{minipage}%
  \hfill
  \begin{minipage}[t]{0.485\linewidth}
    \raggedright
    {\fontsize{9.5}{11}\selectfont\bfseries\texttt{Vidore3EnergyRetrieval}\par}
    \vspace{3pt}
    \vdrExample
      {Contribution of ACER and CRE to the operational implementation of the REMIT regulation}
      {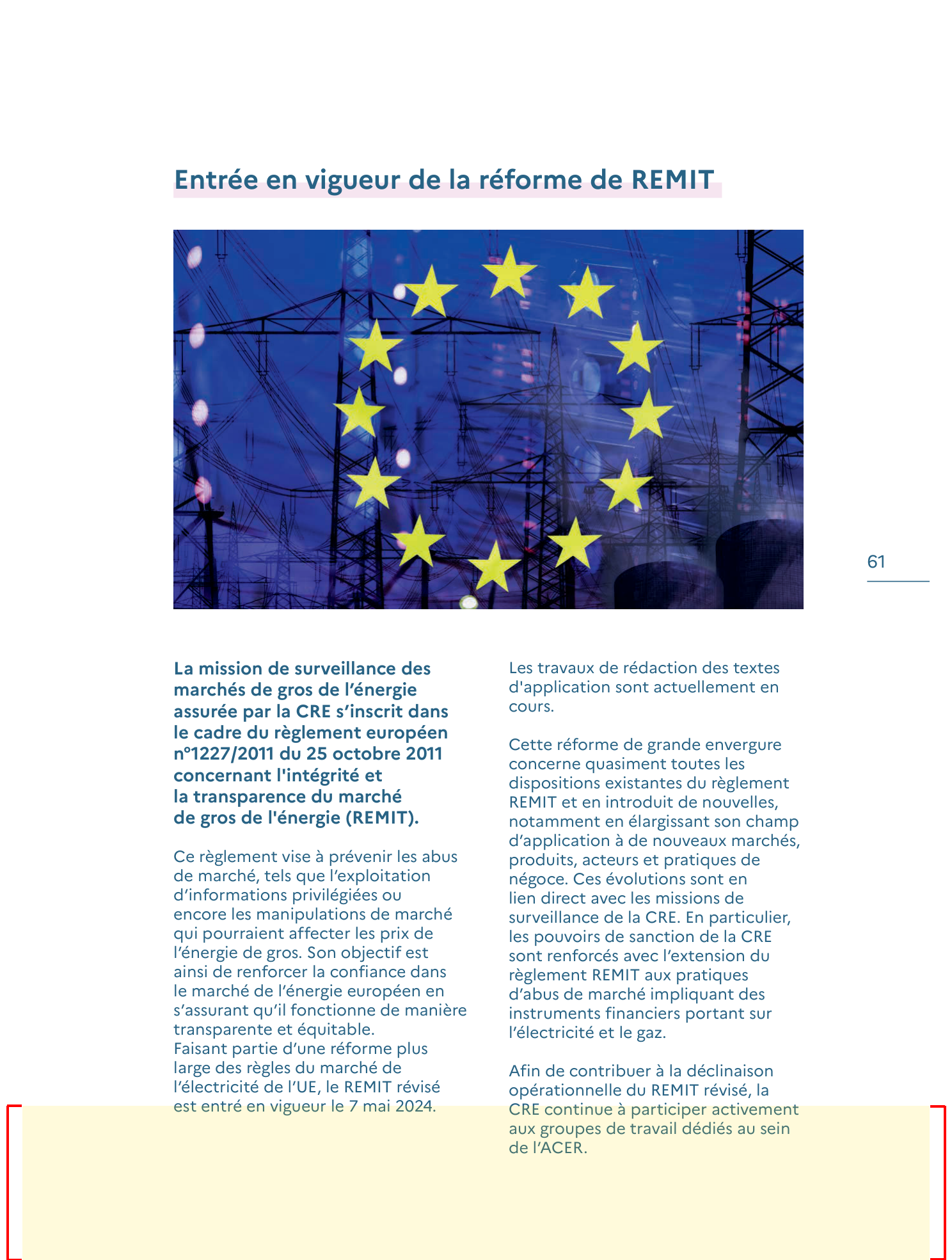}{0bp 0bp 0bp 247bp}
  \end{minipage}%
  \par\vspace{24pt}
  \begin{minipage}[t]{0.485\linewidth}
    \raggedright
    {\fontsize{9.5}{11}\selectfont\bfseries\texttt{Vidore3FinanceEnRetrieval}\par}
    \vspace{3pt}
    \vdrExample
      {Can the depositary for depositary shares be removed without appointing a successor?}
      {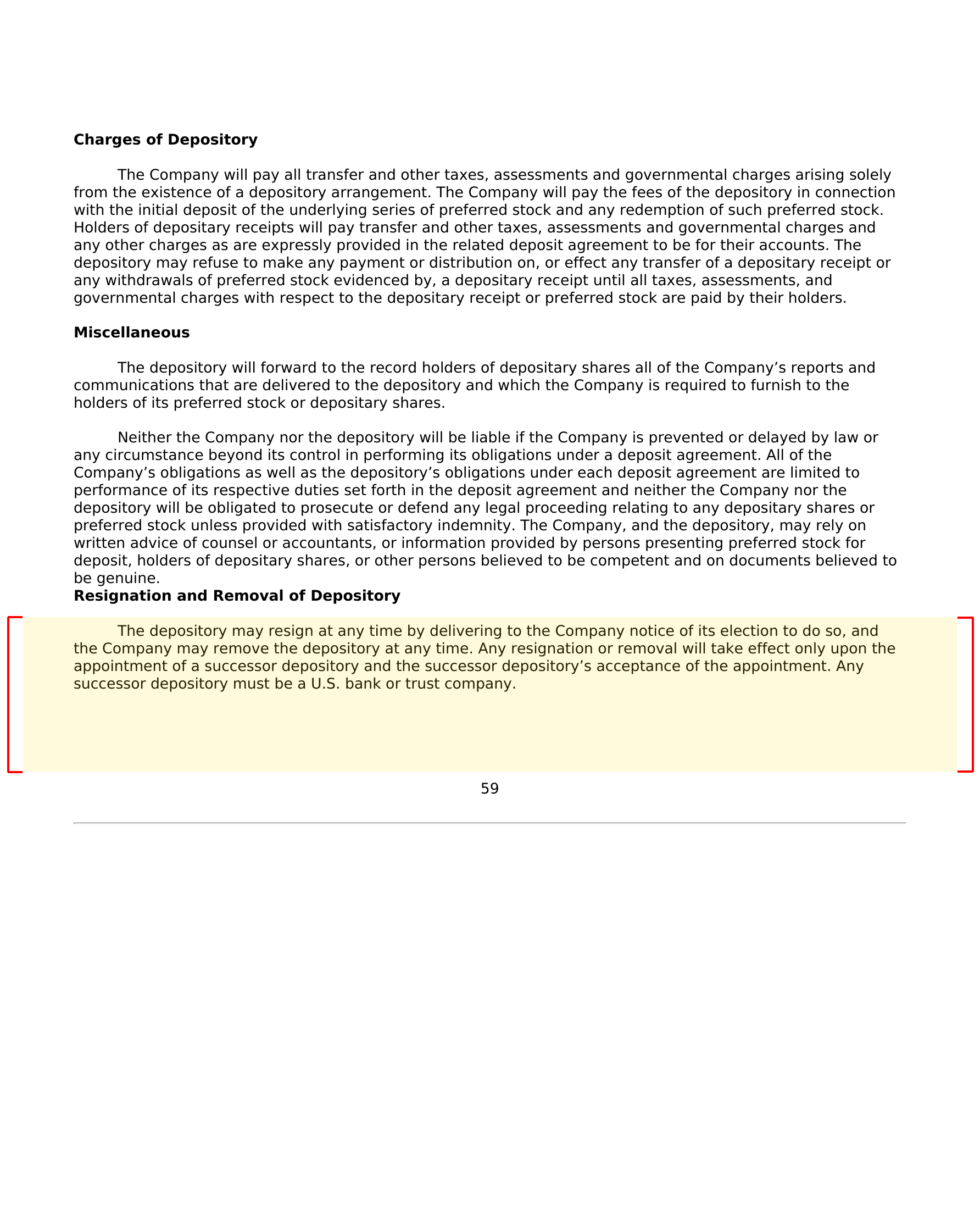}{0bp 684bp 0bp 226bp}
  \end{minipage}%
  \hfill
  \begin{minipage}[t]{0.485\linewidth}
    \raggedright
    {\fontsize{9.5}{11}\selectfont\bfseries\texttt{Vidore3FinanceFrRetrieval}\par}
    \vspace{3pt}
    \vdrExample
      {Indicate the year in which LVMH began participating in the UTTIYAN program.}
      {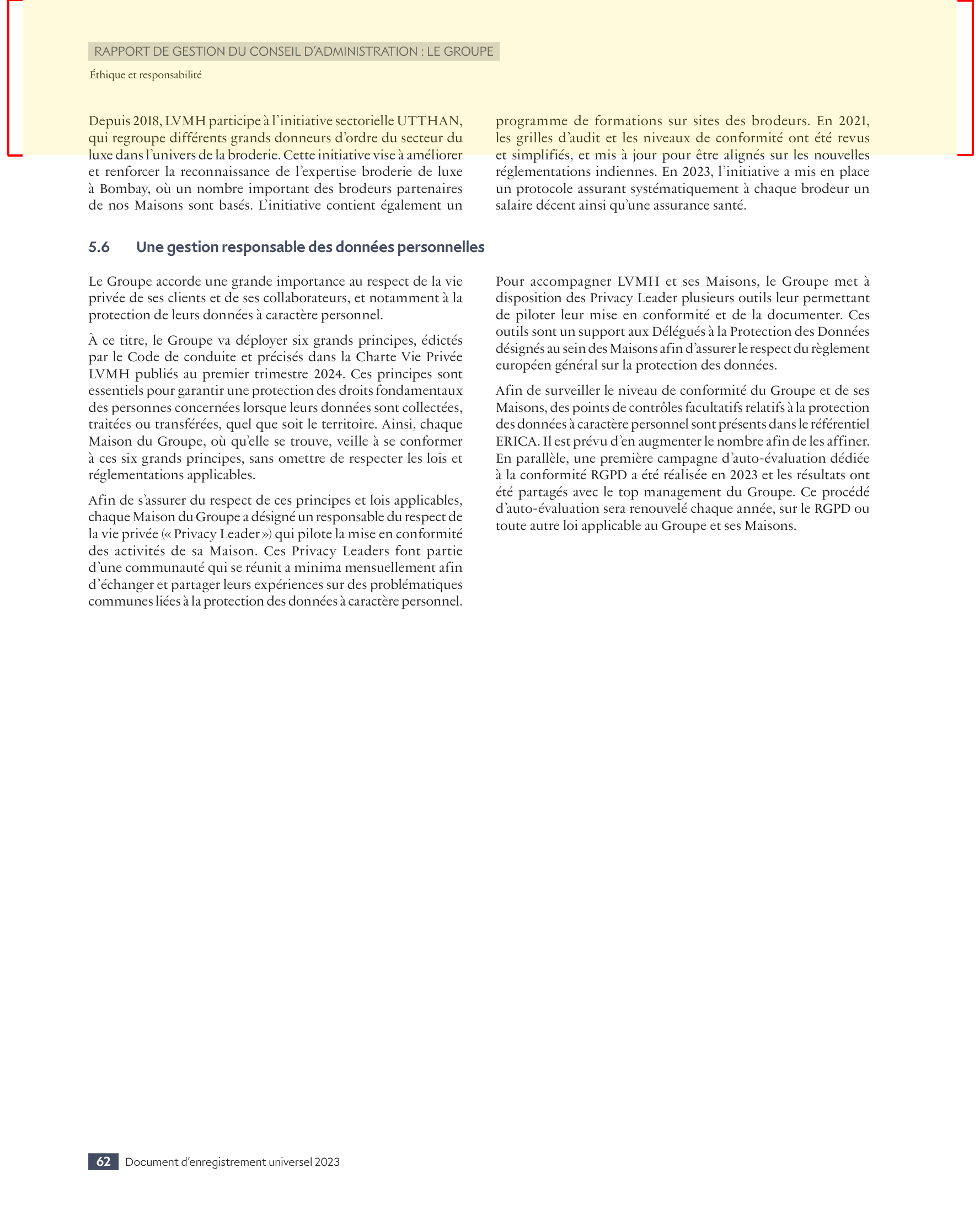}{0bp 67bp 0bp 0bp}
  \end{minipage}%
\end{figure}

\clearpage
\setlength{\vdrImageHeight}{245pt}

\begin{figure}[H]
  \centering
  \begin{minipage}[t]{0.485\linewidth}
    \raggedright
    {\fontsize{9.5}{11}\selectfont\bfseries\texttt{Vidore3HrRetrieval}\par}
    \vspace{3pt}
    \vdrExample
      {What are the targeted policies required to support both workers and economic transformation for EU sustainability?}
      {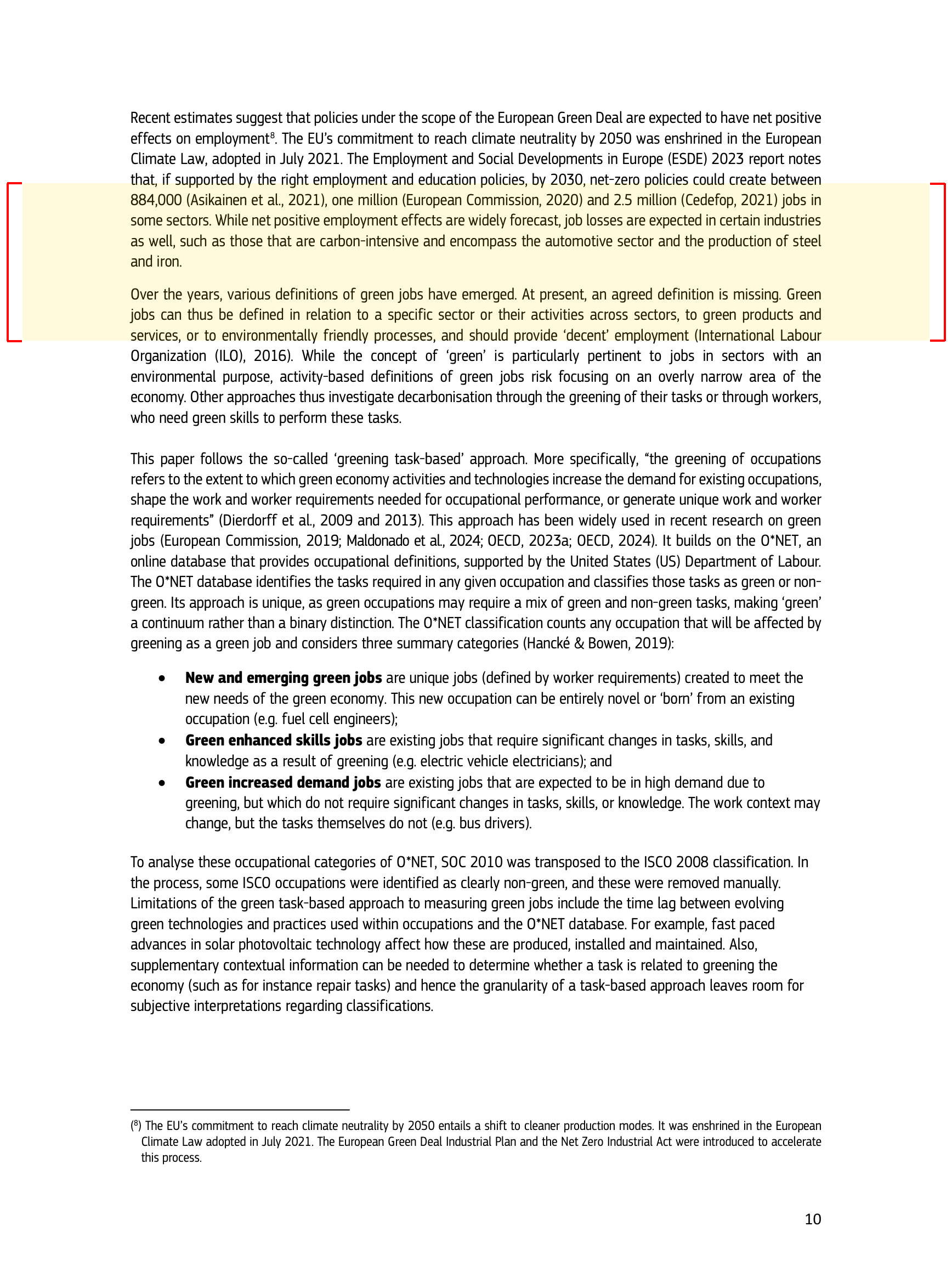}{0bp 90bp 0bp 0bp}
  \end{minipage}%
  \hfill
  \begin{minipage}[t]{0.485\linewidth}
    \raggedright
    {\fontsize{9.5}{11}\selectfont\bfseries\texttt{Vidore3IndustrialRetrieval}\par}
    \vspace{3pt}
    \vdrExample
      {Enumerate the three types of instrumentation that allow for an extended TUS interval, and analyze the maximum allowable working pressure (PSIA) for 1 1/2-inch pipe at Schedule 80 and 160.}
      {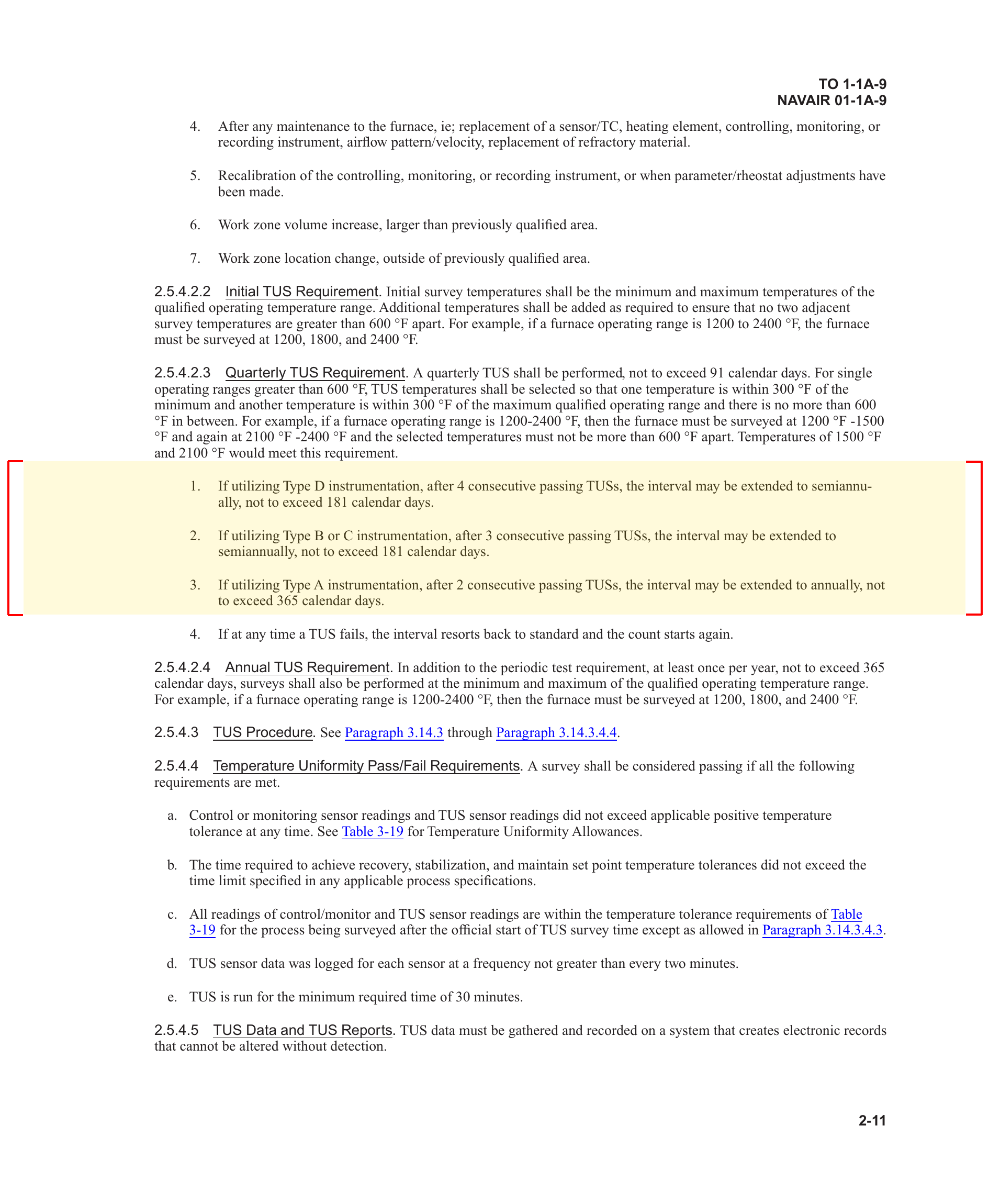}{0bp 134bp 0bp 133bp}
  \end{minipage}%
  \par\vspace{24pt}
  \begin{minipage}[t]{0.485\linewidth}
    \raggedright
    {\fontsize{9.5}{11}\selectfont\bfseries\texttt{Vidore3PharmaceuticalsRetrieval}\par}
    \vspace{3pt}
    \vdrExample
      {How is the FDA using RWD to study the safety and use of CBD, and what are the limitations of this approach?}
      {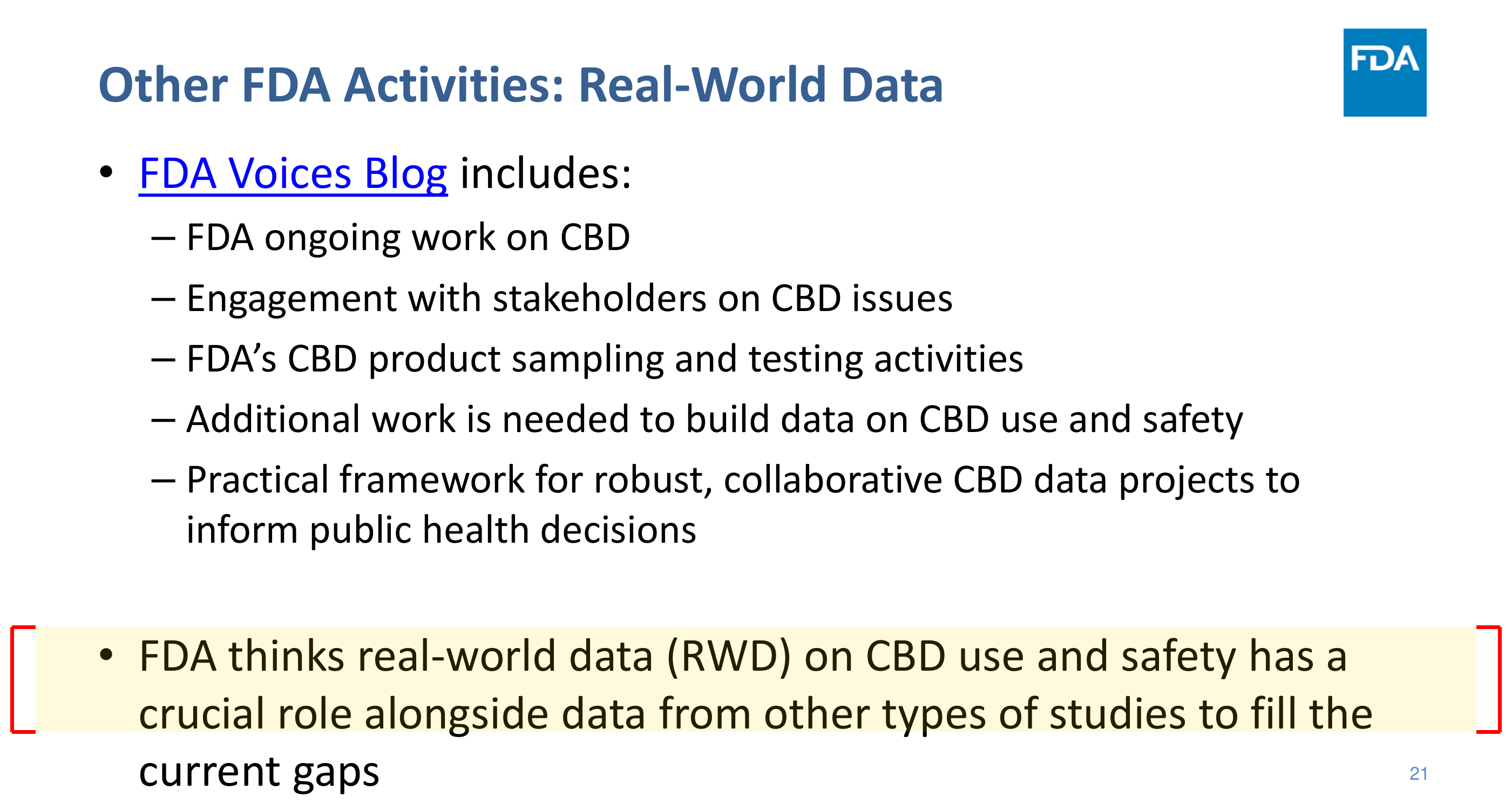}{0bp 13bp 0bp 37bp}
  \end{minipage}%
  \hfill
  \begin{minipage}[t]{0.485\linewidth}
    \raggedright
    {\fontsize{9.5}{11}\selectfont\bfseries\texttt{Vidore3PhysicsRetrieval}\par}
    \vspace{3pt}
    \vdrExample
      {What is the difference between the fractal dimension of cauliflower (approximately 2.33) and that of the Breton coast (approximately 1.25) mentioned in the examples in the document?}
      {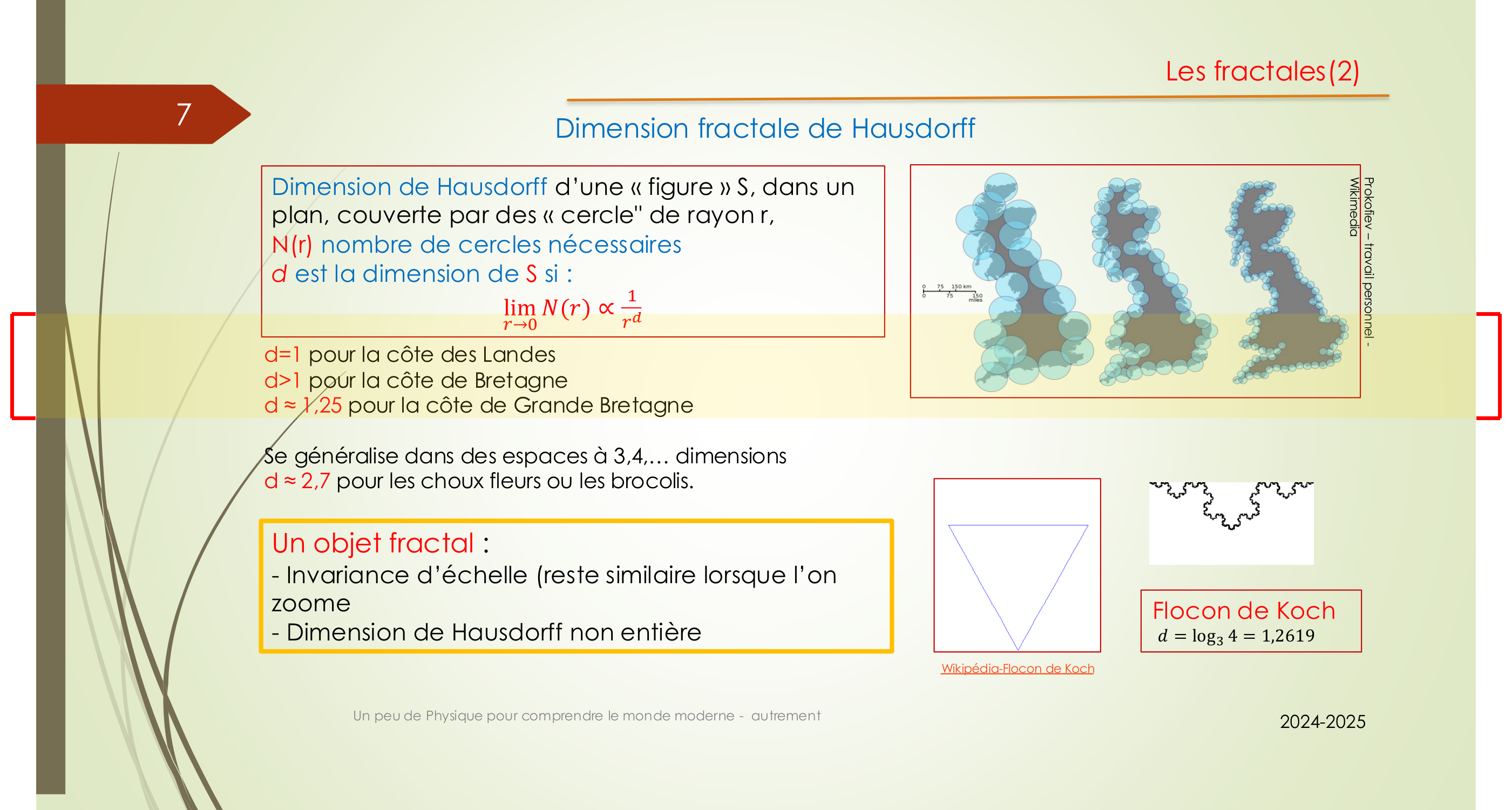}{0bp 0bp 0bp 0bp}
  \end{minipage}%
\end{figure}
\endgroup

\clearpage
\section{Results on BRIGHT}
\label{app:bright}

The BRIGHT \citep{su2025bright} benchmark is one of the few public benchmarks to have chunk-level annotation. However, these chunks are unevenly distributed, such that, for example, in Biology and Earth Science, the average gold chunk size is $9.6\times$ and $25.6\times$ larger than that of negative chunks, respectively.
We therefore re-chunk the corpus using our in-house semantic splitter to a target chunk size of 256 tokens while preserving the original gold boundaries.
We further remove documents containing no negative passages and queries that point to these documents.
Table~\ref{tab:bright-models} shows the results for our models compared against several baselines. We observe that we improve over the v1 baselines across all seven domains, which we attribute to our chunk-level distillation training recipe. 
Our 9B model outperforms voyage-4-large by 6.0 points and trails voyage-context-4 by 0.8 points on average, whereas our 0.6B model trails voyage-4-large by 1.5 points.
BRIGHT queries can require several chunks from several documents, so decomposing or rewriting them can significantly improve performance. In Table~\ref{tab:bright-models}, GPT-4 query reformulations bring no average improvement for our 9B model (0.2 points lower), while the 0.6B model gains 5.2 points.

\begin{table}[H]
    \centering
    \small
    \setlength{\tabcolsep}{4pt}
    \caption{nDCG@10 on re-chunked BRIGHT with original and GPT-4-reformulated queries. Bold marks the best score in each numeric column across all models and query variants. Avg.\ is the macro-average across domains. Scores are in \%. $^{\dagger}$Non-contextual model, which encodes every chunk independently.}
    \label{tab:bright-models}
    \begin{adjustbox}{max width=\textwidth}
    \begin{tabular}{llcccccccc}
      \toprule
      Model & Query & Biology & \shortstack{Earth\\Sci.} & Econ. & Psych. & Robotics & \shortstack{Stack\\Overflow} & \shortstack{Sust.\\Living} & Avg. \\
      \midrule
      \texttt{pplx-embed-v1-4b}$^{\dagger}$ & Org.  & 23.9 & 29.4 & 19.0 & 25.9 & 12.2 & 10.6 & 17.7 & 19.8 \\
                               & GPT-4 & 23.8 & 27.4 & 17.3 & 25.7 & 7.9 & 7.0 & 16.5 & 17.9 \\
      \vonesize{4b} & Org.  & 22.0 & 29.5 & 22.2 & 25.4 & 12.7 & 9.3 & 20.1 & 20.2 \\
                   & GPT-4 & 24.7 & 34.5 & 24.5 & 27.3 & 12.0 & 11.5 & 24.7 & 22.7 \\
      voyage-4-large$^{\dagger}$ & Org.  & 32.7 & 35.6 & 23.9 & 25.5 & 17.9 & 19.5 & 25.9 & 25.9 \\
                     & GPT-4 & 17.6 & 25.2 & 11.9 & 11.9 & 13.2 & 15.8 & 15.2 & 15.8 \\
      voyage-context-4 & Org.  & 38.6 & \textbf{50.2} & \textbf{29.8} & 32.8 & \textbf{27.8} & 18.1 & 31.6 & \textbf{32.7} \\
                       & GPT-4 & 38.0 & 41.3 & 23.7 & 30.5 & 24.6 & 19.9 & 27.8 & 29.4 \\
      \midrule
      \vtwosize{0.6b} & Org.  & 30.0 & 37.5 & 23.4 & 29.0 & 16.2 & 11.6 & 22.9 & 24.4 \\
                  & GPT-4 & \textbf{39.8} & 43.0 & 26.2 & 34.8 & 19.5 & 14.0 & 29.7 & 29.6 \\
      \vtwosize{9b} & Org.  & 34.7 & 42.2 & 27.6 & \textbf{41.1} & 24.4 & 21.2 & \textbf{31.9} & 31.9 \\
                & GPT-4 & 35.5 & 42.6 & 27.2 & 41.0 & 22.4 & \textbf{21.8} & 31.2 & 31.7 \\
      \bottomrule
    \end{tabular}
    \end{adjustbox}
  \end{table}

\clearpage

%% file: figures/visual_chunking.tex
\begingroup
\definecolor{ink}{HTML}{3F464D}
\definecolor{muted}{HTML}{727B85}
\definecolor{bandone}{HTML}{4F93CF}
\definecolor{bandtwo}{HTML}{48A89A}
\definecolor{bandthree}{HTML}{D5AA45}
\definecolor{bandfour}{HTML}{A28AC5}
\definecolor{sepfill}{HTML}{ECEEF1}

\newcommand{\pageimage}{figures/906-Vidore3ComputerScienceRetrieval.png}

\begin{tikzpicture}[
  x=1cm,y=1cm,text=ink,
  font=\fontfamily{phv}\fontsize{8}{10}\selectfont,
  flow/.style={-{Stealth[length=2mm]},draw=ink,line width=.75pt},
  sep/.style={draw=ink!65,fill=sepfill,rounded corners=1pt,
    inner xsep=3pt,inner ysep=2pt,font=\fontfamily{phv}\bfseries\fontsize{6}{7}\selectfont},
  spatial sep/.style={sep,inner xsep=1.5pt,inner ysep=.7pt,
    font=\fontfamily{phv}\bfseries\fontsize{4.5}{5}\selectfont},
  arrow label/.style={inner sep=0pt,font=\fontfamily{phv}\bfseries\fontsize{6.5}{8}\selectfont},
  heading/.style={font=\fontfamily{phv}\bfseries\fontsize{9}{11}\selectfont},
  note/.style={text=muted,font=\fontfamily{phv}\fontsize{7}{9}\selectfont,align=center}
]
\path[use as bounding box] (-.3,1.35) rectangle (17.9,6.15);

\node[heading] at (.95,5.9) {Visual document};
\node[heading] at (5.298,5.9) {Visual patches};
\node[heading] at (9.698,5.9) {Horizontal bands};
\node[heading] at (15.125,5.9) {Soft visual token sequence};

\node[anchor=north west,inner sep=0] at (0,4.97892)
  {\includegraphics[width=1.9cm,height=2.45784cm]{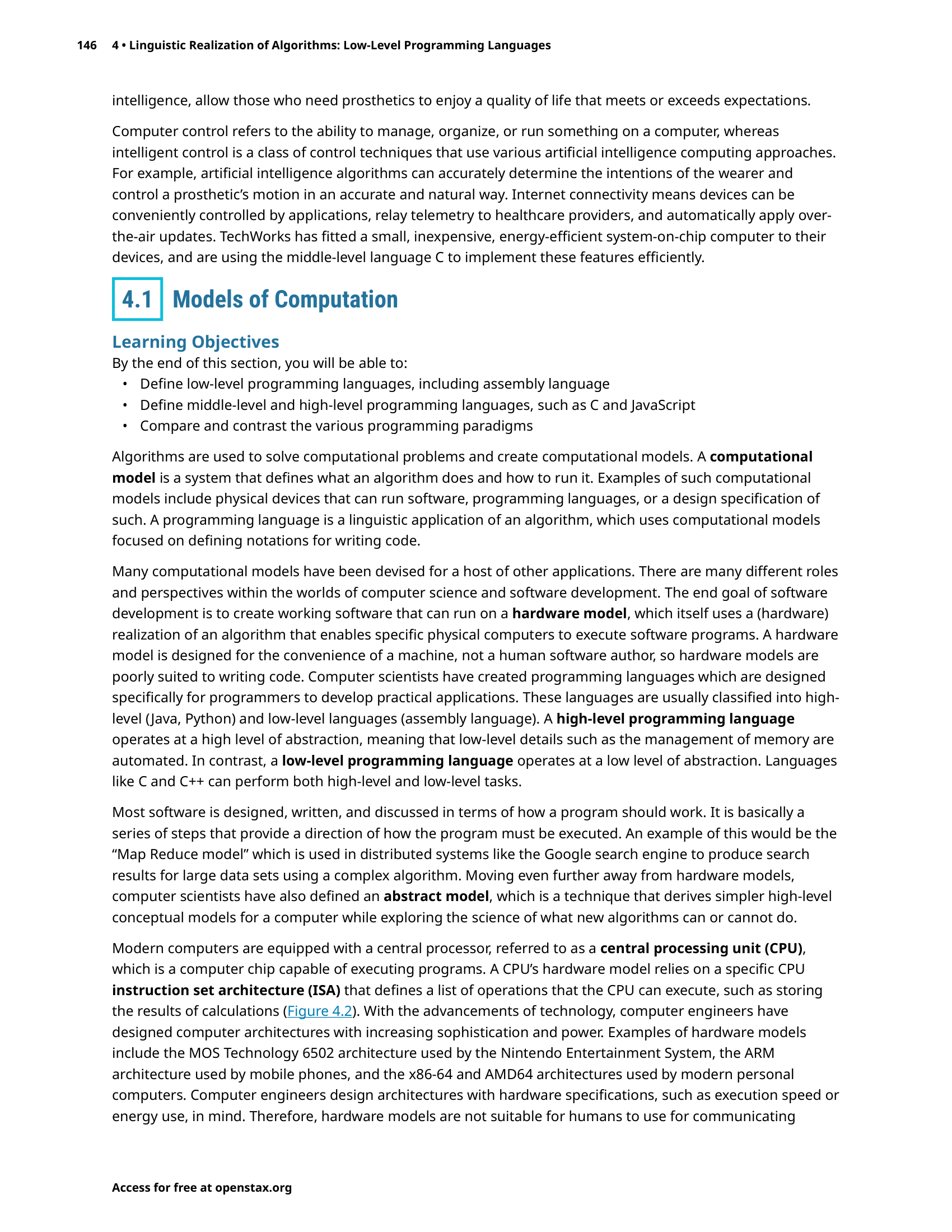}};
\draw[ink!35,line width=.4pt] (0,2.52108) rectangle (1.9,4.97892);
\node[note] at (.95,1.7) {Input page};

\draw[flow] (2.2,3.75) -- ++(1.8,0)
  node[midway,above=3pt,arrow label] {Vision Tower};

\foreach \row in {0,...,11} {
  \foreach \column in {0,...,8} {
    \pgfmathsetmacro{\patchleft}{4.3+\column*(1.9/9+.012)}
    \pgfmathsetmacro{\patchtop}{5.04492-\row*(.20482+.012)}
    \begin{scope}
      \clip (\patchleft,{\patchtop-.20482})
        rectangle ({\patchleft+1.9/9},\patchtop);
      \node[anchor=north west,inner sep=0]
        at ({4.3+\column*.012},{5.04492-\row*.012})
        {\includegraphics[width=1.9cm,height=2.45784cm]{\pageimage}};
      \fill[ink,opacity=.04] (\patchleft,{\patchtop-.20482})
        rectangle ({\patchleft+1.9/9},\patchtop);
    \end{scope}
    \draw[ink!50,line width=.2pt] (\patchleft,{\patchtop-.20482})
      rectangle ({\patchleft+1.9/9},\patchtop);
  }
}
\node[note] at (5.298,1.7) {Spatial token grid};

\draw[flow] (6.6,3.75) -- ++(1.8,0)
  node[midway,above=3pt,arrow label] {Chunk + SEP};

\foreach \i/\col in {0/bandone,1/bandtwo,2/bandthree,3/bandfour} {
  \pgfmathsetmacro{\bandtop}{5.35692-\i*(.63846+.22)}
  \foreach \row in {0,1,2} {
    \foreach \column in {0,...,8} {
      \pgfmathsetmacro{\patchleft}{8.7+\column*(1.9/9+.012)}
      \pgfmathsetmacro{\patchtop}{\bandtop-\row*(.20482+.012)}
      \begin{scope}
        \clip (\patchleft,{\patchtop-.20482})
          rectangle ({\patchleft+1.9/9},\patchtop);
        \node[anchor=north west,inner sep=0]
          at ({8.7+\column*.012},{\bandtop+\i*.61446-\row*.012})
          {\includegraphics[width=1.9cm,height=2.45784cm]{\pageimage}};
        \fill[\col,opacity=.20] (\patchleft,{\patchtop-.20482})
          rectangle ({\patchleft+1.9/9},\patchtop);
      \end{scope}
      \draw[\col!65!ink,line width=.2pt]
        (\patchleft,{\patchtop-.20482}) rectangle ({\patchleft+1.9/9},\patchtop);
    }
  }
}
\foreach \i in {0,1,2} {
  \node[spatial sep] at (9.698,{5.35692-(\i+1)*.63846-(\i+.5)*.22}) {SEP};
}
\node[note] at (9.698,1.7) {Band-aligned patches};

\draw[flow] (11,3.75) -- ++(1.2,0)
  node[midway,above=3pt,arrow label] {Flatten};

\foreach \i/\col in {0/bandone,1/bandtwo,2/bandthree,3/bandfour} {
  \pgfmathsetmacro{\start}{12.5+\i*1.45}
  \foreach \offset in {0,.68} {
    \draw[fill=\col!35,draw=\col!80!ink,line width=.5pt,rounded corners=.5pt]
      ({\start+\offset},3.43) rectangle ++(.22,.64);
  }
  \node[font=\fontfamily{phv}\fontsize{8}{9}\selectfont]
    at ({\start+.45},3.75) {$\cdots$};
  \draw[decorate,decoration={brace,mirror,amplitude=3pt},draw=\col!80!ink]
    (\start,3.23) -- ({\start+.9},3.23);
  \pgfmathtruncatemacro{\bandnumber}{\i+1}
  \node[note,text=\col!65!ink] at ({\start+.45},2.85) {Band \bandnumber};
}
\foreach \i in {0,1,2} {
  \node[sep,minimum width=.41cm,minimum height=.64cm,inner xsep=0pt,
    font=\fontfamily{phv}\bfseries\fontsize{5}{6}\selectfont]
    at ({13.675+\i*1.45},3.75) {SEP};
}
\node[note] at (15.125,1.7) {Row-major tokens with SEP};

\end{tikzpicture}
\endgroup